\documentclass{article}

\usepackage[preprint]{neurips_2024}

\usepackage[utf8]{inputenc}
\usepackage[T1]{fontenc}
\usepackage{hyperref}
\usepackage{url}
\usepackage{booktabs}
\usepackage{amsfonts}
\usepackage{nicefrac}
\usepackage{microtype}
\usepackage{amsmath}
\usepackage{amsthm}
\usepackage{algorithm}
\usepackage{graphicx}
\usepackage{amssymb}
\usepackage{tabularx}
\usepackage[table]{xcolor}
\usepackage{soul}
\usepackage{subcaption}
\usepackage{float}

\usepackage{tikz}
\usetikzlibrary{arrows.meta,positioning}
\usepackage{algorithmic}

\newtheorem{theorem}{Theorem}
\newtheorem{proposition}[theorem]{Proposition}

\newtheorem*{proposition*}{Proposition}

\title{Dynamical Kernel Attention: Estimating the Present from a Record of the Past}

\author{%
  Peter Racioppo \\
  Independent Researcher \\
  Los Angeles, CA \\
  \texttt{pcracioppo@gmail.com} \\
}

\begin{document}

\maketitle

\begin{abstract}
Self-attention encodes where a past token lies, but not how its uncertainty
grows with time: evidence accumulates without bound as the context grows, and
nothing indicates which tokens are safe to discard. We introduce Dynamical
Kernel Attention (DKA), which derives attention from robust estimation of a
latent state governed by a stochastic differential equation. Under this model,
each head draws bounded evidence from any context. This lets each head abstain
when relevant evidence is weak, and gives each head a memory window outside
which discarding tokens changes its output by a provably bounded amount. On
language modeling, DKA outperforms RoPE within the training length and ALiBi
far beyond it, and matches sliding-window attention with a substantially
smaller key–value cache.
\end{abstract}

\section{Introduction}
\label{sec:intro}

Recurrent and state-space models forget through their dynamics: as the state
evolves, information about the past fades. Self-attention has no such mechanism.
Its positional encodings record where a past token is, but not how much it
still tells us about the present, so a distant token can carry as much weight
as a recent one, and models trained on short contexts often fail on longer
ones.

Recency biases \citep{press2022trainshorttestlong} lower the logits of distant
tokens, and learned decays and forget gates
\citep{sun2023retentivenetworksuccessortransformer,lin2025forgetting} make this
discount adaptive. These methods specify how much to discount a token, not why
its information is lost, and because their weights sum to one, a head with no
relevant recent token shifts its weight back to distant ones. Robust Filter
Attention (RFA) \citep{racioppo2026rfa} models the loss of information. It treats
tokens as noisy observations of a latent state evolving under a linear
stochastic differential equation (SDE), so a token's reliability falls as
uncertainty accumulates over the lag. But its flat prior on past states counts
every transported token, however distant, as a full measurement of the present,
so a head's evidence grows without bound with context length.

\begin{figure}[b]
  \vspace{-4mm}
  \centering
  \includegraphics[width=0.9\linewidth]{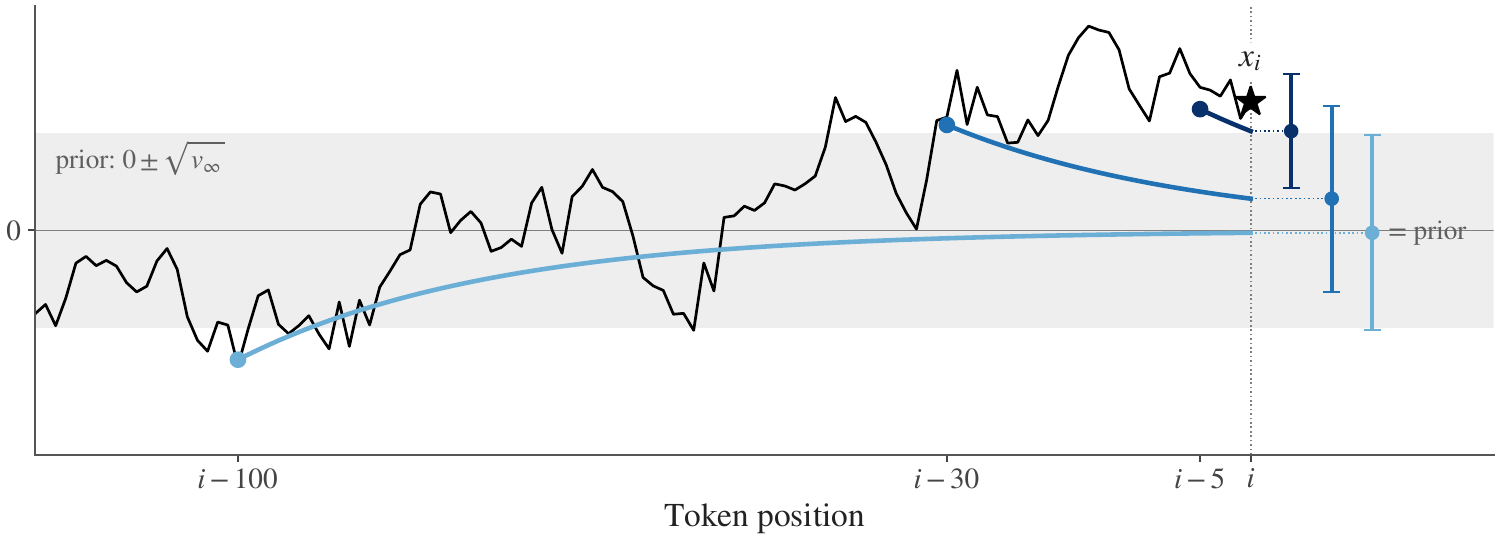}
  \caption{\textbf{A distant token restates the prior.}
  A latent state (black) evolves under stable dynamics around an equilibrium
  at zero; the star marks the current state $x_i$. Past tokens are carried
  forward to the present (blue), with error bars showing prediction uncertainty.
  A nearby token (dark) predicts $x_i$ closely; a distant one (light) decays
  to the equilibrium, with uncertainty $\sqrt{v_\infty}$.}
  \label{fig:intro_schematic}
\end{figure}

We introduce Dynamical Kernel Attention
(DKA)\footnote{Code: \url{https://github.com/PCR-git/Dynamical-Kernel-Attention}.}, in which past states are drawn
from the stationary distribution of the process. Under stable dynamics, a
distant token's prediction has decayed to equilibrium and says nothing about
the present, so its evidence vanishes (Figure~\ref{fig:intro_schematic}).
Attention then counts the information tokens carry, and each head's evidence
is bounded, as a recurrent model's memory is. The current estimate enters the
normalization as a null slot, so a head with weak evidence abstains instead of
redistributing its weight. Since each key's weight is capped by its lag,
whatever its content, each head has a memory window set by its decay rate,
beyond which evicting tokens changes its output by a bounded amount, for any input. Our contributions are:

\textbf{(i)} A stationary measurement model where information inherently fades with lag, ensuring each head's total evidence remains strictly bounded in infinite contexts (Proposition~\ref{prop:evidence}).

\textbf{(ii)} An attention update derived as a robust estimation step. The prior produces a null slot in the normalization, allowing heads with weak evidence to abstain.

\textbf{(iii)} A per-head memory window set by the decay rate, guaranteeing that distant tokens in the KV-cache can be evicted with a strictly bounded change to the output (Proposition~\ref{prop:window}).

\textbf{(iv)} A probabilistic interpretation of standard positional encodings, separating deterministic state transport (RoPE) from latent measurement uncertainty and information decay (ALiBi and KERPLE).

\section{Related Work}
\label{sec:RelatedWork}

\paragraph{Probabilistic and kernel views of attention.}
Attention has been interpreted as inference in probabilistic memories
\citep{gabbur2021probabilistic}, Gaussian mixtures
\citep{nguyen2022improving}, associative energy models
\citep{ramsauer2021hopfieldnetworksneed}, Gaussian processes
\citep{bui2025revisitingkernelattentioncorrelated}, Bayesian positional priors
\citep{bianchessi2026bayesian}, and Nadaraya--Watson regression
\citep{tsai2019transformerdissectionunifiedunderstanding}. These generally use static similarity or covariance. DKA instead derives its kernel from latent dynamics: transport, uncertainty, and information decay emerge analytically from a shared stochastic state model.

\paragraph{Attention and state estimation.}
Standard self-attention theoretically represents Kalman filters via kernel smoothing analogies \citep{goel2024transformerrepresentkalmanfilter}, and frozen Transformers implicitly execute optimal filtering via in-context learning on dynamical trajectories \citep{akram2026transformers}. While Kalman Linear Attention parameterizes state-space models using parallelized information-form Kalman updates \citep{shaj2026kalmanlinearattentionparallel}, it compresses history into a hidden state. DKA instead keeps explicit pairwise interactions with past tokens. DKA keeps the model of RFA \citep{racioppo2026rfa} but conditions observations on the current state under a stationary process, and treats each layer as a proximal robust estimation step; together, these bound each head's evidence and memory.

\paragraph{Positional encoding and extrapolation.}
Rotation \citep{su2023roformerenhancedtransformerrotary} encodes where a past
token lies; distance penalties such as ALiBi \citep{press2022trainshorttestlong}
encode how its reliability falls with lag, without RoPE's fine-grained
localization. xPos and RetNet combine rotation with multiplicative decay
\citep{sun2022lengthextrapolatabletransformer,sun2023retentivenetworksuccessortransformer},
and RoPE is usually extended by rescaling its frequencies
\citep{peng2023yarnefficientcontextwindow,ding2024longrope}. In DKA, as in RFA
\citep{racioppo2026rfa}, both follow from one dynamical model
(Appendix~\ref{app:pos_encodings}): transport gives RoPE's rotation, and the
information each key carries gives ALiBi's bias, with slope set by the decay.
The null slot plays the role of attention sinks \citep{xiao2024attentionsinks}
and ``softmax plus one'' \citep{miller2023offbyone}, pulling the update toward
no change.

\paragraph{Cache eviction.}
To prune the key--value cache, most methods track received attention \citep{zhang2023h2o,li2024snapkv}. Others size the cache per head empirically \citep{xiao2025duoattention}. RazorAttention sets each ALiBi head's window from its slope via a per-key worst-case score bound computed from weight norms \citep{tang2025razorattention}. DKA's measurement noise model natively bounds the total evidence from distant tokens (Proposition~\ref{prop:evidence}), capping every key's weight and yielding a principled memory window without computing score bounds.


\section{Method}
\label{sec:method}

Each DKA layer takes one step of a robust estimator of a latent state. Past
tokens, transported to the current position by the dynamics, become
measurements of the current state whose information fades with lag; the layer
reweights them at the current estimate and steps toward the estimate they
support.

\subsection{Model}
\label{sec:model}

We model each token embedding $\boldsymbol z_j\in\mathbb R^d$ as a noisy observation of a latent state $\boldsymbol x(t)\in\mathbb C^d$ evolving under a linear SDE:
\begin{equation}
\label{eq:SDE}
d\boldsymbol x(t) = \boldsymbol A\boldsymbol x(t)\,dt
    + \boldsymbol G\,d\boldsymbol W(t), \qquad
\boldsymbol z_j = \boldsymbol C\bigl(\boldsymbol x(t_j)
    + \boldsymbol\varepsilon_j\bigr),
\end{equation}
where $\boldsymbol W$ is a Wiener process,
$\boldsymbol Q=\boldsymbol G\boldsymbol G^\dagger$,
$\boldsymbol\varepsilon_j\sim\mathcal N(\boldsymbol 0,\boldsymbol R)$,
$\boldsymbol A$ is Hurwitz, and $\boldsymbol C$ is invertible. We take
$\boldsymbol A$, $\boldsymbol Q$, and $\boldsymbol R$ to be diagonal, with
the eigenvalues of $\boldsymbol A$ in conjugate pairs so that
$\boldsymbol z_j$ is real. This loses no generality when the system matrices
share an eigenbasis, which can be absorbed into $\boldsymbol C$.

Attention relates pairs of positions, so we express the state at $t_i$ in
terms of the state at an earlier $t_j$. Writing
$\boldsymbol x_i:=\boldsymbol x(t_i)$, $\Delta t_{ij}=t_i-t_j$, and
$\boldsymbol\Phi_{ij}=e^{\boldsymbol A\Delta t_{ij}}$, integrating
Eq.~\ref{eq:SDE} over $[t_j,t_i]$ gives
\begin{equation}
\label{eq:transition}
\boldsymbol x_i=\boldsymbol\Phi_{ij}\boldsymbol x_j+\boldsymbol F_{ij},
\qquad
\boldsymbol F_{ij}:=\int_{t_j}^{t_i}e^{\boldsymbol A(t_i-\tau)}\boldsymbol G\,d\boldsymbol W(\tau),
\end{equation}
the transported past state plus the stochastic forcing accumulated since. Since
$\boldsymbol F_{ij}$ depends only on the noise after $t_j$, it is independent
of $\boldsymbol x_j$, and it is Gaussian,
$\boldsymbol F_{ij}\sim\mathcal N(\boldsymbol 0,\boldsymbol V_{ij})$, with
$\boldsymbol V_{ij}:=\boldsymbol V(\Delta t_{ij})$ the solution of the differential Lyapunov equation (DLE),
\begin{equation}
\label{eq:DLE}
\dot{\boldsymbol V}(s)=\boldsymbol A\boldsymbol V(s)+\boldsymbol V(s)\boldsymbol A^\dagger+\boldsymbol Q,
\qquad
\boldsymbol V(0)=\boldsymbol 0 .
\end{equation}


\subsection{Likelihood}
\label{sec:factors}

While a Kalman filter recursively compresses history, self-attention maintains explicit pairwise access to all past tokens. To bridge this, each token $\boldsymbol z_j$ supplies a state estimate $\hat{\boldsymbol x}_j=\boldsymbol C^{-1}\boldsymbol z_j=\boldsymbol x_j+\boldsymbol\varepsilon_j$, transported independently to position $i$ via deterministic dynamics: $\hat{\boldsymbol x}_{ij}:=\boldsymbol\Phi_{ij}\hat{\boldsymbol x}_j$ for $j\le i$. Fusing this ensemble exactly requires their joint distribution, as they correlate through shared stochastic forcing. We instead use a composite likelihood \citep{lindsay1988composite}, the product of
the individual factors:
\begin{equation}
\label{eq:composite}
p\big(\{\hat{\boldsymbol x}_{ij}\}_{j\le i}\mid\boldsymbol x_i\big)
\approx\prod_{j\le i}p(\hat{\boldsymbol x}_{ij}\mid\boldsymbol x_i).
\end{equation}
This requires only that each factor be correctly specified
\citep{varin2011overview}. It also matches the pairwise structure of
attention: each transported estimate is compared with the current state on its own, whereas under the exact likelihood, newer tokens supersede the older evidence they overlap with, as in a Kalman filter, whether or not they are relevant. We now compute the factors $p(\hat{\boldsymbol x}_{ij}\mid\boldsymbol x_i)$.

Each transported estimate is the current state plus an error,
\[
\hat{\boldsymbol x}_{ij}
=\boldsymbol\Phi_{ij}\boldsymbol x_j+\boldsymbol\Phi_{ij}\boldsymbol\varepsilon_j
=\boldsymbol x_i-\boldsymbol F_{ij}+\boldsymbol\Phi_{ij}\boldsymbol\varepsilon_j .
\]
Since $\boldsymbol x_i$ contains $\boldsymbol F_{ij}$, each factor depends on
the joint distribution of $\boldsymbol F_{ij}$ and $\boldsymbol x_i$, and
hence on the prior on the past state. Under
$\boldsymbol x_j\sim\mathcal N(\boldsymbol 0,\boldsymbol\Pi)$, this pair is
jointly Gaussian with
$\operatorname{Cov}(\boldsymbol F_{ij},\boldsymbol x_i)=\boldsymbol V_{ij}$ and
$\operatorname{Var}(\boldsymbol x_i)=\boldsymbol M_{ij}:=\boldsymbol\Phi_{ij}\boldsymbol\Pi\boldsymbol\Phi_{ij}^\dagger+\boldsymbol V_{ij}$,
so
$\mathbb E[\boldsymbol F_{ij}\mid\boldsymbol x_i]=\boldsymbol V_{ij}\boldsymbol M_{ij}^{-1}\boldsymbol x_i$
and
$\operatorname{Var}(\boldsymbol F_{ij}\mid\boldsymbol x_i)=\boldsymbol V_{ij}-\boldsymbol V_{ij}\boldsymbol M_{ij}^{-1}\boldsymbol V_{ij}=\boldsymbol H_{ij}\boldsymbol V_{ij}$,
with $\boldsymbol H_{ij}:=\boldsymbol I-\boldsymbol V_{ij}\boldsymbol M_{ij}^{-1}$.
Hence,
\begin{equation}
\label{eq:general_factor}
\hat{\boldsymbol x}_{ij}\mid\boldsymbol x_i\sim
\mathcal N\big(\boldsymbol H_{ij}\boldsymbol x_i,\,\boldsymbol\Sigma_{ij}\big),
\qquad
\boldsymbol\Sigma_{ij}=\boldsymbol H_{ij}\boldsymbol V_{ij}+\boldsymbol\Phi_{ij}\boldsymbol R\boldsymbol\Phi_{ij}^\dagger .
\end{equation}
Each past token is a linear measurement of the current state, as in a Kalman
update, with observation matrix $\boldsymbol H_{ij}$, the fraction of
$\boldsymbol x_i$ not attributed to forcing.

\paragraph{Diffuse prior (RFA).}
Assuming no prior knowledge of the past state corresponds to a diffuse prior ($\boldsymbol\Pi\to\infty$). Under this limit, $\boldsymbol M_{ij}^{-1}\to\boldsymbol 0$, so $\boldsymbol H_{ij}\to\boldsymbol I$ and
$\boldsymbol\Sigma_{ij}\to\boldsymbol V_{ij}+\boldsymbol\Phi_{ij}\boldsymbol R\boldsymbol\Phi_{ij}^\dagger$,
the measurement model of RFA \citep{racioppo2026rfa}. Every token, however
distant, is then an unbiased measurement of $\boldsymbol x_i$ whose variance
tends to $\boldsymbol V_\infty:=\lim_{s\to\infty}\boldsymbol V(s)$, so its
information is at least $\boldsymbol V_\infty^{-1}$ and the total grows
linearly with context length.

\paragraph{Stationary prior (DKA).}
We instead take the past state to be a draw from the stationary
distribution of the process, $\mathcal N(\boldsymbol 0,\boldsymbol V_\infty)$,
the standard initialization for a stable component of a state-space model
\citep{durbin_koopman_2012time}, which a stable process approaches from any
initial condition. Since this distribution
is preserved by the dynamics, $\operatorname{Var}(\boldsymbol x_i)=\boldsymbol V_\infty$,
and taking the variance of Eq.~\ref{eq:transition} gives
$\boldsymbol\Phi_{ij}\boldsymbol V_\infty\boldsymbol\Phi_{ij}^\dagger
=\boldsymbol V_\infty-\boldsymbol V_{ij}$. Hence
$\boldsymbol M_{ij}=\boldsymbol V_\infty$, and Eq.~\ref{eq:general_factor}
holds with
\begin{equation}
\label{eq:factor}
\boldsymbol H_{ij}=\boldsymbol I-\boldsymbol V_{ij}\boldsymbol V_\infty^{-1}.
\end{equation}
The likelihood of the current state is therefore
\begin{equation}
\label{eq:likelihood}
p\big(\{\hat{\boldsymbol x}_{ij}\}_{j\le i}\mid\boldsymbol x_i\big)
\approx\prod_{j\le i}\mathcal N\big(\hat{\boldsymbol x}_{ij};\,\boldsymbol H_{ij}\boldsymbol x_i,\,\boldsymbol\Sigma_{ij}\big).
\end{equation}
At zero lag, $\boldsymbol H_{ii}=\boldsymbol I$ and
$\boldsymbol\Sigma_{ii}=\boldsymbol R$, and the two priors nearly agree
while $\boldsymbol H_{ij}\approx\boldsymbol I$. As the lag grows,
$\boldsymbol V_{ij}\to\boldsymbol V_\infty$ and $\boldsymbol H_{ij}\to\boldsymbol 0$,
so the information
$\boldsymbol H_{ij}^\dagger\boldsymbol\Sigma_{ij}^{-1}\boldsymbol H_{ij}$
that a token carries about $\boldsymbol x_i$ vanishes, and the overcount of
shared forcing is bounded in context length (Proposition~\ref{prop:evidence}). The diffuse prior instead erroneously counts a distant token as a full measurement of the present, ignoring that a stable process forgets its past.



\paragraph{Robust factors.}
The Gaussian factor assumes the realized forcing has covariance
$\boldsymbol V_{ij}$. To allow larger deviations, and either endpoint to be
atypical, we give each pair a latent scale
$s_{ij}\sim\mathrm{Inv\text{-}Gamma}(\nu/2,\nu/2)$ on its covariance,
\[
\hat{\boldsymbol x}_{ij}\mid\boldsymbol x_i,s_{ij}\sim
\mathcal N\big(\boldsymbol H_{ij}\boldsymbol x_i,\,s_{ij}\boldsymbol\Sigma_{ij}\big).
\]
Integrating out the scale gives a Student-$t$ factor with $\nu$ degrees of
freedom \citep{Lange1989RobustSM},
\begin{equation}
\label{eq:robust_likelihood}
p(\hat{\boldsymbol x}_{ij}\mid\boldsymbol x)\;\propto\;
|\boldsymbol\Sigma_{ij}|^{-1/2}\Big(1+\frac{d_{ij}^2(\boldsymbol x)}{\nu}\Big)^{-(\nu+d)/2},
\end{equation}
where
$d_{ij}^2(\boldsymbol x)=(\hat{\boldsymbol x}_{ij}-\boldsymbol H_{ij}\boldsymbol x)^\dagger\boldsymbol P_{ij}(\hat{\boldsymbol x}_{ij}-\boldsymbol H_{ij}\boldsymbol x)$
and $\boldsymbol P_{ij}:=\boldsymbol\Sigma_{ij}^{-1}$. As $\nu\to\infty$, the
scale concentrates at one and the factor becomes Gaussian.

\subsection{Robust Estimation Step}
\label{sec:weights}
\label{sec:map}

Each layer refines an existing current-state estimate $\hat{\boldsymbol x}_i$,
used as the prior $\boldsymbol x_i\sim\mathcal N(\hat{\boldsymbol x}_i,\boldsymbol U)$
with diagonal covariance $\boldsymbol U$. We weight each measurement by its precision, the information it carries,
times a robust weight,
\begin{equation}
\label{eq:weights}
w_{ij}=\Big(1+\frac{d_{ij}^2}{\nu}\Big)^{-\kappa},
\qquad
\kappa=\frac{\nu+d}{d},
\end{equation}
which is one at perfect agreement and falls as the measurement disagrees with
the current estimate. Any weight that falls with the residual defines a
robust estimator \citep{huber1964}; this one is tied to the noise model: for
an isotropic factor,
$\boldsymbol P_{ij}w_{ij}\propto p(\hat{\boldsymbol x}_{ij}\mid\boldsymbol x)^{2/d}$,
the factor's likelihood per dimension. The power $2/d$ keeps the weight on a
fixed scale as the dimension grows. A consistent token has
$d_{ij}^2\approx d$, so its likelihood is about $e^{-d/2}$ times that of a
perfectly agreeing one, and in high dimension the weights would be nearly
all-or-nothing. With the power $2/d$, and $\nu$ proportional to $d$
(Section~\ref{sec:dka}), the typical weight no longer depends on $d$, analagous to how the
$1/\sqrt d$ scaling keeps attention logits on a fixed scale. As
$\nu\to\infty$, the weight tends to $\exp(-d_{ij}^2/d)$.

Each layer takes one step toward the minimizer of
\begin{equation}
\label{eq:neglogpost}
\mathcal L(\boldsymbol x)
=\tfrac12(\boldsymbol x-\hat{\boldsymbol x}_i)^\dagger\boldsymbol U^{-1}(\boldsymbol x-\hat{\boldsymbol x}_i)
+\tfrac12\sum_{j\le i}\rho\big(d_{ij}^2(\boldsymbol x)\big),
\qquad
\rho(u)=d\Big[1-\Big(1+\frac{u}{\nu}\Big)^{-\nu/d}\Big],
\end{equation}
whose penalty $\rho$ has the weight as its derivative. Setting the gradient of
$\mathcal L$ to zero gives
\[
\boldsymbol U^{-1}(\boldsymbol x-\hat{\boldsymbol x}_i)
=\sum_{j\le i}w_{ij}(\boldsymbol x)\,\boldsymbol H_{ij}^\dagger\boldsymbol P_{ij}
(\hat{\boldsymbol x}_{ij}-\boldsymbol H_{ij}\boldsymbol x),
\]
which is implicit, since the weights depend on the solution. Holding the
weights fixed at the current estimate, where they depend on the residual
$\boldsymbol r_{ij}:=\hat{\boldsymbol x}_{ij}-\boldsymbol H_{ij}\hat{\boldsymbol x}_i$
through $d_{ij}^2=\boldsymbol r_{ij}^\dagger\boldsymbol P_{ij}\boldsymbol r_{ij}$,
makes it linear. Since $\rho$ is concave, this minimizes a quadratic upper
bound on $\mathcal L$ centered at $\hat{\boldsymbol x}_i$: a
majorize--minimize step, which cannot increase $\mathcal L$
\citep{hunterlange2004}. Solving the linear equation, with
$\boldsymbol J_i:=\sum_{j\le i}w_{ij}\boldsymbol H_{ij}^\dagger\boldsymbol P_{ij}\boldsymbol H_{ij}$,
gives the update as a correction to the current estimate,
\begin{equation}
\label{eq:update}
\bar{\boldsymbol x}_i
=\hat{\boldsymbol x}_i+\big(\boldsymbol U^{-1}+\boldsymbol J_i\big)^{-1}
\sum_{j\le i}w_{ij}\boldsymbol H_{ij}^\dagger\boldsymbol P_{ij}\boldsymbol{r}_{ij}.
\end{equation}
This is an information-form Kalman measurement update applied to a batch of
transported measurements. While recursive filters absorb one measurement per
step to avoid dense covariance inversions, our composite likelihood allows
parallel updates from all past tokens. This preserves self-attention's
pairwise routing at the cost of the bounded overcount in
Proposition~\ref{prop:evidence}.

\subsection{Isotropic Implementation}
\label{sec:dka}

We learn separate projections
$\boldsymbol W_q,\boldsymbol W_k,\boldsymbol W_v\in\mathbb C^{d\times d}$, each
absorbing its own $\boldsymbol C^{-1}$, and an output projection
$\boldsymbol W_o$ absorbing $\boldsymbol C$. The current estimate is read as
the query $\boldsymbol Q_i$ and past estimates as keys $\boldsymbol K_j$ when
computing weights, and both through the value projection $\boldsymbol V$ when
aggregating; under a single observation model, the three projections
coincide.

\paragraph{Isotropic decay and noise.}
To retain the memory cost of standard attention, we impose per head
$\boldsymbol A=-\mu\boldsymbol I+i\boldsymbol\Omega$, with
$\boldsymbol\Omega=\mathrm{diag}(\omega_k)$, and take the forcing,
observation-noise, and prior covariances to be $\sigma^2\boldsymbol I$,
$\eta^2\boldsymbol I$, and $\gamma^2\boldsymbol I$. We write
$\boldsymbol E[i,j]=e^{-\mu\Delta t_{ij}}$ for the decay over a lag. Since
$\boldsymbol\Omega$ is real, $\boldsymbol A+\boldsymbol A^\dagger=-2\mu\boldsymbol I$,
so the rotation drops out of Eq.~\ref{eq:DLE}, whose solution is
$\boldsymbol V(s)=v_\infty(1-e^{-2\mu s})\boldsymbol I$ with
$v_\infty=\sigma^2/2\mu$. Hence $\boldsymbol V_{ij}=v_\infty(1-\boldsymbol E[i,j]^2)\boldsymbol I$
and $\boldsymbol V_\infty=v_\infty\boldsymbol I$, and with
$\boldsymbol\Phi_{ij}\boldsymbol\Phi_{ij}^\dagger=\boldsymbol E[i,j]^2\boldsymbol I$,
Eqs.~\ref{eq:general_factor} and~\ref{eq:factor} reduce to scalars:
\begin{equation}
\label{eq:iso_cov}
h_{ij}=\boldsymbol E[i,j]^2,
\qquad
\boldsymbol\Sigma_{ij}=\boldsymbol E[i,j]^2\,\boldsymbol\Sigma_n[i,j]\,\boldsymbol I,
\qquad
\boldsymbol\Sigma_n:=v_\infty(1-\boldsymbol E^2)+\eta^2 .
\end{equation}
The observation coefficient is the fraction of the stationary variance not yet
replaced by forcing, and depends only on the decay rate. Zero-decay heads are
the limit $\mu\to0$ with $\sigma^2$ fixed, in which $h\equiv1$ and
$\boldsymbol V(s)\to\sigma^2s\,\boldsymbol I$, so
$v_\infty(1-\boldsymbol E^2)\to\sigma^2\Delta t$.

\paragraph{Rotated residual.}
Reading the current estimate as the query and past estimates as keys, the
residual of Section~\ref{sec:weights} is
$\boldsymbol r_{ij}=\boldsymbol\Phi_{ij}\boldsymbol K_j-h_{ij}\boldsymbol Q_i$.
With $\tilde{\boldsymbol\Phi}^\pm[k,i]=e^{\pm i\omega_k t_i}$, the transport of
mode $k$ factors as
$\boldsymbol E[i,j]\,\tilde{\boldsymbol\Phi}^+[k,i]\,\tilde{\boldsymbol\Phi}^-[k,j]$.
Rotating the queries, keys, and values,
$\tilde{\boldsymbol Q}=\tilde{\boldsymbol\Phi}^-\odot\boldsymbol Q$,
$\tilde{\boldsymbol K}=\tilde{\boldsymbol\Phi}^-\odot\boldsymbol K$,
$\tilde{\boldsymbol V}=\tilde{\boldsymbol\Phi}^-\odot\boldsymbol V$, and using
$h_{ij}=\boldsymbol E[i,j]^2$ and
$\boldsymbol Q_i=\tilde{\boldsymbol\Phi}^+_i\odot\tilde{\boldsymbol Q}_i$,
\[
\boldsymbol r_{ij}=\boldsymbol E[i,j]\,\tilde{\boldsymbol\Phi}^+_i\odot\boldsymbol R_{qk}[i,j],
\qquad
\boldsymbol R_{qk}[i,j]:=\tilde{\boldsymbol K}_j-\boldsymbol E[i,j]\tilde{\boldsymbol Q}_i .
\]
The residual can thus be formed at the key's own time, where it compares the
key with $\boldsymbol E[i,j]\tilde{\boldsymbol Q}_i$, the past state predicted
by the present, $\mathbb E[\boldsymbol x_j\mid\boldsymbol x_i]$, in the rotated
frame. The rotation has unit magnitude, so
$\|\boldsymbol r_{ij}\|^2=\boldsymbol E[i,j]^2\|\boldsymbol R_{qk}[i,j]\|^2$,
and with $\boldsymbol\Sigma_{ij}=\boldsymbol E[i,j]^2\boldsymbol\Sigma_n[i,j]\boldsymbol I$
the factor $\boldsymbol E^2$ cancels between the residual and its covariance:
\begin{equation}
\label{eq:residual_norm}
\boldsymbol D^2=\|\boldsymbol R_{qk}\|^2\oslash\boldsymbol\Sigma_n,
\qquad
\|\boldsymbol R_{qk}[i,j]\|^2
=\|\boldsymbol K_j\|^2+\boldsymbol E[i,j]^2\|\boldsymbol Q_i\|^2
-2\boldsymbol E[i,j]\,\mathrm{Re}\big(\tilde{\boldsymbol Q}_i^\dagger\tilde{\boldsymbol K}_j\big).
\end{equation}

\paragraph{Logits and value update.}
Setting $\nu=\nu_s d$, so that $\kappa=\nu_s+1$, the combined attention weight
$a_{ij}:=w_{ij}^{\tau}/\boldsymbol\Sigma_n[i,j] = \exp(\boldsymbol L[i,j])$
with logits
\begin{equation}
\label{eq:logits}
\boldsymbol L=-\log\boldsymbol\Sigma_n
-\tau\,(\nu_s+1)\log\!\Big(1+\tfrac{1}{\nu_sd}\boldsymbol D^2\Big)+\boldsymbol M_\le ,
\end{equation}
where $\boldsymbol M_\le$ is the causal mask, including the diagonal, and $\tau>0$ is a learned inverse temperature. In the value basis, the residual
is
$\boldsymbol r^v_{ij}=\boldsymbol E[i,j]\tilde{\boldsymbol\Phi}^+_i\odot\tilde{\boldsymbol V}_j-h_{ij}\boldsymbol V_i$,
and since $\tilde{\boldsymbol\Phi}^+_i$ does not depend on $j$, it factors out
of the weighted sum. The update is therefore the innovation
\begin{equation}
\label{eq:dka_update}
\Delta\boldsymbol V_i
=\frac{\tilde{\boldsymbol\Phi}^+_i\odot\sum_{j\le i}a_{ij}\boldsymbol E[i,j]\tilde{\boldsymbol V}_j-S_i\boldsymbol V_i}{\gamma^{-2}+S_i},
\qquad
S_i:=\sum_{j\le i}a_{ij}h_{ij}.
\end{equation}
Writing
$\bar{\boldsymbol V}_i:=\tilde{\boldsymbol\Phi}^+_i\odot\sum_{j\le i}a_{ij}\boldsymbol E[i,j]\tilde{\boldsymbol V}_j/S_i$
for the estimate the context supports, the update can also be written as
$\Delta\boldsymbol V_i=\frac{S_i}{\gamma^{-2}+S_i}\big(\bar{\boldsymbol V}_i-\boldsymbol V_i\big)$.

\paragraph{Residual connection.}
To update the state estimate without altering the residual stream itself, we apply the step as a correction. The layer adds the mapped innovation, $\boldsymbol Z^+=\boldsymbol Z+\boldsymbol W_o\Delta\boldsymbol V$, taking a partial step toward the context's estimate while the prior state passes through unchanged.

\paragraph{Spectrum.}
To suppress phase wrap-around at long lags, the decay derived by the model must be coupled to the rotation frequency. The heads therefore split one geometric bank of frequencies, $\omega_k=\text{base}^{-k/128}$, and each head's decay rate is tied to its highest frequency, $\mu_h=b\,\omega_{h,\max}$ \citep{racioppo2026rfa}. A head's weight then falls as $e^{-b\theta}$ while its fastest mode turns through $\theta$ radians, so decay is measured per turn rather than per token. The heads thus act as one filter at different time scales: a constant-Q filter bank, as in wavelet analysis.

\paragraph{Bounded update.}
Two differences from softmax attention bound the update. The robust weight is
at most one, so $a_{ij}\le1/\boldsymbol\Sigma_n[i,j]$: no content match can
exceed what its lag allows. The prior adds $\gamma^{-2}$ to the denominator of
Eq.~\ref{eq:dka_update}, a null slot with logit $-\log\gamma^2$ and zero
innovation, so a head with weak evidence makes little update, and the
remaining keys are not amplified when others carry no weight or are evicted.

\paragraph{Recovery of positional encodings.}
A DKA head contains RoPE and ALiBi as special cases (Appendix~\ref{app:pos_encodings}). Its rotation $\tilde{\boldsymbol\Phi}^-$ is RoPE's; with zero decay and no forcing, it reduces to RoPE attention up to value rotation. With decay, writing $\pi_{ij}:=a_{ij}h_{ij}/S_i$, the estimate is
\[
\bar{\boldsymbol V}_i=\sum_{j\le i}\pi_{ij}\,
\frac{\tilde{\boldsymbol\Phi}^+_i\odot\tilde{\boldsymbol V}_j}{\boldsymbol E[i,j]},
\qquad
\pi_{ij}\propto\exp\big(\boldsymbol L[i,j]-2\mu\Delta t_{ij}\big),
\]
which matches ALiBi \citep{press2022trainshorttestlong} with slope $2\mu_h$, applied to transported values rescaled by $1/\boldsymbol E$. Rescaling corrects for transport shrinking each value toward equilibrium by $\boldsymbol E$. Weighting values directly gives ALiBi's exact form but shrinks the estimate (D12, Section~\ref{sec:exp_ablations}). The bias is the logarithm of the key's information $h_{ij}=\boldsymbol E[i,j]^2$, yielding a linear penalty at every lag set by the decay rate. RFA's diffuse prior sets $h\equiv1$, lacking this term; its lag bias $-\log\boldsymbol\Sigma$ is linear only near zero lag for zero-decay heads \citep{racioppo2026rfa}. Two further limits exist: with zero decay, the update is a softmax over innovations with a null logit (``softmax plus one'' \citep{miller2023offbyone}); and since the spectrum ties frequencies to decay, a head lacking decay and forcing is attention without positional encoding (NoPE) \citep{kazemnejad2023impact}.

\subsection{Evidence, Bias, and Memory}
\label{sec:theory}

A model of forgetting should ensure that the evidence a head draws does not
grow without bound, that stale tokens do not bias its estimate, and that
tokens can be discarded at a known cost. We state these properties for a
single head with decay $\mu>0$ and memory length
$\ell:=1/(1-e^{-2\mu})\approx1/(2\mu)$.

\begin{proposition}[Evidence and bias; informal]
\label{prop:evidence}
For data drawn from the model, with unit robust weights, the information DKA
assigns to any number of recent tokens lies between the exact information and
$\ell/\eta^2$, and its estimate retains all of the current state on average.
The information RFA assigns grows linearly with the number of tokens $N$, and
its estimate retains a fraction of order $\ell/N$ of the current state
(Appendix~\ref{app:counting_evidence}).
\end{proposition}

The composite likelihood can only overcount, and under the stationary prior
the overcount is capped, so a head's evidence, and with it its step size,
converges as the context grows. Under the diffuse prior, every distant token
restates the stationary distribution, so only about $\ell$ of RFA's tokens
carry the current state, and the rest pull its estimate toward the
equilibrium.

\begin{proposition}[Memory window]
\label{prop:window}
If every value has norm at most $B$, discarding the keys at lag $T$ or more
changes a head's update by at most $B\,\theta(3+\theta)\,e^{-\mu T}$, with
$\theta=\gamma^2/\big(\eta^2(1-e^{-\mu})\big)$, for every query, context
length, and choice of content weights (Appendix~\ref{app:memory_window}).
\end{proposition}

A head therefore needs only the keys at lag below
$\mu^{-1}\log\big(\theta(3+\theta)/\varepsilon\big)$ to stay within
$\varepsilon B$ of its full-context output, so its cache can be sized from its
parameters before seeing any input. The bound uses both the cap
$a_{ij}\le1/\boldsymbol\Sigma_n$ and the null slot; RFA has no such window,
because a distant token's weight in its normalizer is not decayed.


\section{Experiments}
\label{sec:experiments}

We test three claims: DKA trained on 512 tokens stays stable to 8192 tokens
with no change at inference, and is better than standard attention within the
training length; each head's cache can be pruned to a window set from its
learned parameters; and both results require the null slot and DKA's bounded
weights, not only the decay bias its spectrum implies.

\subsection{Setup}
\label{sec:exp_setup}

The main experiments and the ablations train on WikiText-103 \citep{merity2016pointersentinelmixturemodels} with 6-layer, 19.4M-parameter decoder-only models with 8 heads per layer, on 512-token contexts.
As in RFA, every
model, including the baselines, works in a complex space of dimension 32 per
head, so parameter counts match exactly (Appendix~\ref{app:training}). We
report test perplexity on non-overlapping blocks of 512 to 8192 tokens, with
no fine-tuning. Results are from one training run per model. Comparisons
between models are paired over test blocks, and we report the percentage
change in perplexity with its standard error over blocks. This error reflects
the finite test set; with one run per model, it does not include variation
between training runs.

\paragraph{Spectrum.}
DKA, its ablations and RFA use RoPE's frequencies unchanged, with base
$10^4$, and only add the coupled decay $\mu_h=b\,\omega_{h,\max}$
(Section~\ref{sec:dka}). We use $b=1$: each head turns one radian within its
horizon, and the horizons are 3162, 1000, 316, 100, 32, 10, 3 and 1 tokens
(Appendix~\ref{app:spectrum} compares other values of $b$).

\paragraph{Baselines.}
B1, RoPE \citep{su2023roformerenhancedtransformerrotary} at the standard base
$10^4$, also with YaRN at inference \citep{peng2023yarnefficientcontextwindow}
(without its usual fine-tune) and with a 512-key sliding window at inference;
B2, ALiBi \citep{press2022trainshorttestlong}; and B3, spectrally coupled RFA
\citep{racioppo2026rfa} on the same spectrum as DKA.

\paragraph{Ablations.}
D1--D13 each change one part of DKA; D14--D17 go step by step from
DKA to rotary attention with ALiBi slopes derived from its spectrum; D18--D23
add DKA's components to the baselines (Appendix~\ref{app:training}).


\subsection{Length Extrapolation}
\label{sec:exp_extrapolation}

\begin{table}[b]
\centering
\caption{Test perplexity on WikiText-103 at increasing context lengths, for models trained on 512 tokens.
Perplexity covers every position of each block. B1 + YaRN, B1 + window: RoPE with YaRN or with a 512-key sliding
window at inference.}
\label{tab:extrapolation}
\small
\begin{tabular}{lccccc}
\toprule
Model & 512 & 1024 & 2048 & 4096 & 8192 \\
\midrule
B1, RoPE                   & 27.93 & 40.55 & 100.63 & 202.36 & 305.69 \\
B1 + YaRN                  & 27.93 & 27.70 & 29.37  & 32.97  & 38.08 \\
B1 + window                & 27.93 & 26.79 & 26.18  & 26.07  & 25.67 \\
B2, ALiBi                  & 28.63 & 27.33 & 26.57  & 26.41  & 25.99 \\
B3, RFA                    & 27.74 & 26.47 & 26.01  & 27.15  & 30.02 \\
\midrule
DKA                        & \textbf{27.59} & \textbf{26.39} & \textbf{25.75} & \textbf{25.64} & \textbf{25.25} \\
\bottomrule
\end{tabular}
\end{table}

DKA is better than every baseline at every length (Table~\ref{tab:extrapolation}),
and its perplexity keeps falling to 8192 tokens, 16 times the training length.
Within the training length, it beats RoPE by $1.2\pm0.2\%$, RFA on the same
spectrum by $0.5\pm0.2\%$, and ALiBi by $3.7\pm0.2\%$ (paired over 547 test
blocks). RoPE at the standard base fails beyond 512 tokens; YaRN, applied without
fine-tuning, recovers most of the loss but still degrades with length (38.08
at 8192 tokens). RFA on the same spectrum degrades beyond 2048 tokens: the
penalty on its distant tokens levels off, so their share of the weight grows
with the context (Appendix~\ref{app:rfa_sink}).

A 512-key sliding window also removes RoPE's failure and is better than ALiBi
at every length, so it is the strongest practical baseline; DKA is better than
it at every length (25.25 against 25.67 at 8192 tokens). On positions 512--4096, DKA with full context beats it by
$1.8\pm0.3\%$: about $1.2\%$ from better modeling within the window and
$0.6\pm0.1\%$ from context beyond it (DKA against its own 512-key window).
ALiBi gains more from distant context ($1.4\%$) but stays $3.0\pm0.3\%$ behind
DKA at 4096 tokens. The
gain beyond the window comes from the two slowest heads, mostly in one layer
(Appendix~\ref{app:spectrum}).

\subsection{Per-Head Cache Eviction}
\label{sec:exp_eviction}

We evaluate on positions 2048--4096 of 4096-token test blocks, beyond the
training length and where the slow heads' windows are cut, and report the
mean number of keys kept per head. A \emph{sliding window} keeps the $W$ most
recent keys in every head; a \emph{calibrated} window keeps, per head, the
keys at lag below the point where the head's measured weight becomes
negligible on held-out data; and a \emph{bound} window keeps the keys at lag
below the window implied by Proposition~\ref{prop:window}, which uses only
learned parameters (Appendix~\ref{app:training}).

\begin{figure}[bt]
\centering
\includegraphics[width=0.8\linewidth]{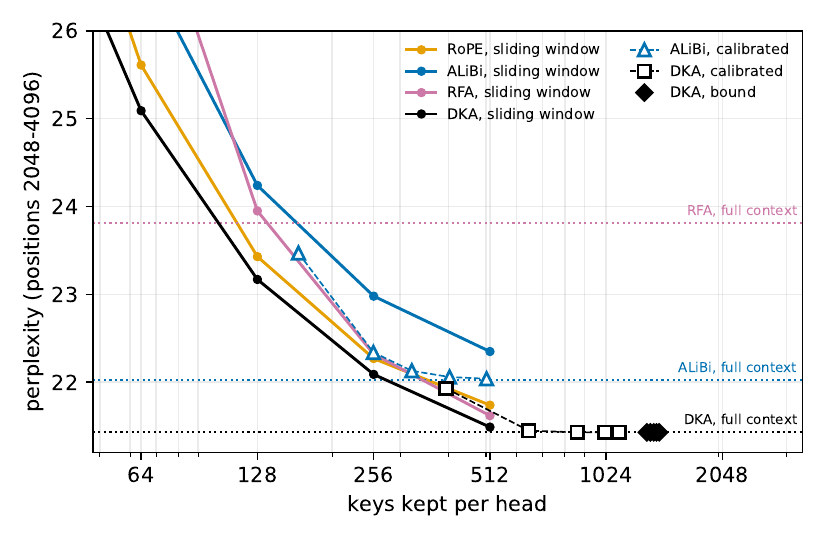}
\caption{Perplexity against cache size at 4096 tokens (positions 2048--4096): sliding windows (lines),
calibrated windows ($\varepsilon$ from 0.1 to 0.001, open markers), DKA's bound windows ($\varepsilon$ from 0.3 to
0.003, filled diamonds), ALiBi's slope windows ($\varepsilon$ from 0.5 to 0.001, triangles), DKA's calibrated and bound
windows capped at 512 keys (crosses) and each model's full context (dotted). RoPE's full context (396.8) and RFA's calibrated
windows (40.5--63.0) are off the axis, as are those of D17, rotary attention with the bias of DKA's decay
(28.40--29.92, full context 28.58).}
\label{fig:eviction}
\end{figure}

Five of DKA's eight heads have horizons of at most 100 tokens and need only a
few recent keys (Figure~\ref{fig:eviction}; Table~\ref{tab:eviction} in
Appendix~\ref{app:eviction}). Windows set by the bound alone keep 1382 keys per
head, 45\% of the cache, and match DKA's full-context perplexity with a
worst-case error of $0.004$; windows calibrated on validation data keep 862
keys at the same perplexity (worst-case error $0.12$).

Capping each head's window at the training length leaves 324 keys per head, a
cache 9.5 times smaller than full context, and matches DKA's 512-key sliding
window (21.49); a sliding window of the same size is worse (21.83). At the same
cache size, DKA is better than ALiBi with per-head windows (21.49 at 324 keys
against 22.13 at 321), and better than ALiBi with full context (22.03). RFA
cannot be pruned head by head: its distant keys form an implicit sink
(Appendix~\ref{app:rfa_sink}).

\subsection{Ablations}
\label{sec:exp_ablations}

Table~\ref{tab:ablation} (Appendix~\ref{app:ablations}) lists every ablation.
The null slot is part of what makes DKA extrapolate: without it (D8), perplexity at
4096 tokens rises from 25.64 to 32.36, while within the training length
nothing changes ($-0.2\pm0.2\%$). Its calibrated cache windows also become
unsafe (worst-case error 1.72, against 0.12).

D17 keeps only DKA's spectrum and the decay
bias it implies, which makes it rotary attention with ALiBi slopes $2\mu_h$.
It reaches 31.02 at 4096 tokens and 33.45 at 8192, against DKA's 25.64 and
25.25, and is worse within the window ($+1.0\pm0.2\%$). Within DKA, scoring
keys by the dot product (D4), which drops the floor of the residual, drifts
beyond 2048 tokens (36.05 at 8192). In DKA, a key's weight is at most its
value in agreement, so the decay bounds what a distant key can contribute; a
dot-product score has no such bound, and in the slowest heads, whose bias is
only $-2.6$ at lag 4096, a strong distant match can outweigh it. The bound also
makes the cache windows safe: with calibrated windows, the worst-case error is
2.71 for D17 and 2.32 for D4, against 0.12 for DKA, and D17 keeps more keys
(928 against 862). DKA's null slot does not rescue a dot-product score: added
to D17 with the innovation output (D16), it makes the drift worse (50.97 at
4096 tokens, against 31.02).

Finally, DKA weights each key's estimate of the current state rather than its
value. Weighting the values as in softmax attention (D14) costs about 1\% at
every length but keeps the extrapolation, with a smaller and safer cache (710
keys per head, worst-case error 0.03), so it improves perplexity without being
what makes DKA stable (Appendix~\ref{app:ablations}).

The rescaling by $1/E$ is safe only because the match is attenuated. In the
residual $\|\tilde K_j-E_{ij}\tilde Q_i\|^2$, a distant key matches only if it
has shrunk by $E_{ij}$, which offsets the factor $1/E_{ij}$ its estimate
receives. A dot-product score without the attenuation does not test this, so
a distant key with a strong match can take weight up to $1/E_{ij}$. Three
ablations isolate the two factors. With DKA's values, the unattenuated score
(D5) diverges (56564 at 2048 tokens). The same score with the values weighted
as in softmax attention (D15), which differs from D5 only in the rescaling,
drifts but does not diverge (54.81 at 2048 tokens). Restoring the attenuation
with DKA's values (D4), which differs from D5 only in the match, brings it
back to a mild drift (26.26 at 2048 tokens, 36.05 at 8192).



\section{Conclusion}
\label{sec:conclusion}

We reformulate self-attention as robust iterative estimation of a latent state evolving under stochastic dynamics. Conditioning transported observations on the
current state makes the information a distant token carries fade with lag, so each head draws bounded evidence, abstains through a null slot when that
evidence is weak, and has a memory window beyond which tokens can be evicted with a guaranteed bound on the change to its output. On language modeling, DKA outperforms RoPE within the training length and ALiBi far beyond it, and matches sliding-window attention with a substantially smaller cache.

DKA assumes a stationary process, which fits transients far from equilibrium poorly. Since a head's memory window depends strictly on lag, tokens outside the slowest horizons cannot be retrieved regardless of content match. The composite likelihood also overcounts shared forcing, making learned noise parameters effective rather than physical. Future work could generalize the latent space to hyperspherical geometry, where directional query-key matching recovers the standard dot product and avoids value rescaling. Additionally, pooling tokens exactly by accounting for shared forcing could improve estimation where evidence is heavily correlated.

\bibliographystyle{plainnat}
\bibliography{references}

\newpage

\appendix


\section*{Appendix Contents}

\newcommand{\appentry}[2]{%
  \noindent\textbf{Appendix~\ref{#1}: \nameref{#1}}\hfill\pageref{#1}\\
  \hspace*{1em}\textit{#2}\par\medskip}

\appentry{app:implementation}{Real-valued implementation and numerical
stability, with pseudocode and a block diagram for a single isotropic head.}

\appentry{app:pos_encodings}{RoPE, ALiBi, NoPE and softmax plus one as limits of a DKA head, including a derivation of ALiBi's linear distance bias.}

\appentry{app:counting_evidence}{Formal statement and proof of
Proposition~\ref{prop:evidence}: how much evidence a head draws compared with
the exact amount, and the bias of its estimate.}

\appentry{app:memory_window}{Proof of Proposition~\ref{prop:window}: each head
can discard tokens beyond a window set by its decay rate, with a bounded change
to its output.}

\appentry{app:training}{Model and training settings, baselines and
ablations, the spectrum, and the evaluation and eviction protocols.}

\appentry{app:additional_experiments}{Perplexity by position, the spectrum
sweep, and further ablation and eviction results.}


\newpage

\section{Implementation of the DKA Layer}
\label{app:implementation}

DKA is formulated in a complex latent
space, $\mathbb C^d\cong\mathbb R^{2d}$, and implemented with real operations, following \citep{racioppo2026rfa}. The realization follows RFA's, except for the residual, the
information $S_i$, and the proximal update.

\paragraph{Projections.}
A complex linear map $\boldsymbol W=\boldsymbol W_r+i\boldsymbol W_i$ with
bias $\boldsymbol b=\boldsymbol b_r+i\boldsymbol b_i$ acts on paired real and
imaginary channels. Since the inputs and outputs of the head are real, only
part of it is needed:
\[
\mathcal L^{d\to2d}(\boldsymbol x)
=\begin{bmatrix}\boldsymbol W_r\\\boldsymbol W_i\end{bmatrix}\boldsymbol x
+\begin{bmatrix}\boldsymbol b_r\\\boldsymbol b_i\end{bmatrix},
\qquad
\mathcal L^{2d\to d}(\boldsymbol y)
=\begin{bmatrix}\boldsymbol W_r&-\boldsymbol W_i\end{bmatrix}\boldsymbol y
+\boldsymbol b_r,
\]
both standard real linear layers. The projections
$\boldsymbol Q,\boldsymbol K,\boldsymbol V=\mathcal L_{q,k,v}^{d\to2d}(\boldsymbol Z)$
are split into real and imaginary halves, $\boldsymbol Q=[\boldsymbol Q_{\mathrm{re}};\boldsymbol Q_{\mathrm{im}}]$.
We do not constrain paired coordinates to be conjugate; the real output
projection plays the role of taking the real part.

\paragraph{Rotation.}
With $\boldsymbol C_\omega[k,i]=\cos(\omega_kt_i)$ and
$\boldsymbol S_\omega[k,i]=\sin(\omega_kt_i)$, multiplication by
$\tilde{\boldsymbol\Phi}^-$ is
\[
\tilde{\boldsymbol Q}_{\mathrm{re}}=\boldsymbol C_\omega\odot\boldsymbol Q_{\mathrm{re}}+\boldsymbol S_\omega\odot\boldsymbol Q_{\mathrm{im}},
\qquad
\tilde{\boldsymbol Q}_{\mathrm{im}}=\boldsymbol C_\omega\odot\boldsymbol Q_{\mathrm{im}}-\boldsymbol S_\omega\odot\boldsymbol Q_{\mathrm{re}},
\]
and likewise for $\boldsymbol K$ and $\boldsymbol V$, which is RoPE's
rotation. The frequencies come in conjugate pairs,
$\boldsymbol\omega=(\omega_1,-\omega_1,\dots,\omega_{d/2},-\omega_{d/2})$,
so that the dynamics are real, and only $d/2$ of them are specified.

\paragraph{Residual and logits.}
Each term of the key-frame residual
(line~7 of Algorithm~\ref{alg:dka}) is real. The squared norms
$\|\boldsymbol Q_i\|^2$ and $\|\boldsymbol K_j\|^2$ are taken in
$\mathbb R^{2d}$ and are unchanged by rotation, so they are computed from the
unrotated projections, and
$\mathrm{Re}\big(\tilde{\boldsymbol Q}^\dagger\tilde{\boldsymbol K}\big)
=\tilde{\boldsymbol Q}_{\mathrm{re}}^\top\tilde{\boldsymbol K}_{\mathrm{re}}+\tilde{\boldsymbol Q}_{\mathrm{im}}^\top\tilde{\boldsymbol K}_{\mathrm{im}}$
is the ordinary inner product of the stacked rotated vectors. With
$\boldsymbol G:=\tilde{\boldsymbol Q}^\top\tilde{\boldsymbol K}\in\mathbb R^{N\times N}$,
a single real matrix product,
\[
\|\boldsymbol R_{qk}[i,j]\|^2
=\|\boldsymbol K_j\|^2+\boldsymbol E[i,j]^2\|\boldsymbol Q_i\|^2
-2\,\boldsymbol E[i,j]\,\boldsymbol G[i,j].
\]
The covariance $\boldsymbol\Sigma_n$, the logits $\boldsymbol L$ and the
weights $\boldsymbol A[i,j]=\tilde a_{ij}=e^{\boldsymbol L[i,j]-m_i}$ are
then elementwise functions of real $N\times N$ matrices. As in RFA, we write
$\hat{\boldsymbol A}:=\boldsymbol A\odot\boldsymbol E$ for the decayed
weights.

\paragraph{Stabilized weights.}
As a softmax does internally, we subtract the row maximum, including the null
slot's logit $-\log\gamma^2$:
$m_i=\max\big(\max_j\boldsymbol L[i,j],\,-\log\gamma^2\big)$.
Multiplying numerator and denominator of Eq.~\ref{eq:dka_update} by $e^{-m_i}$
and writing $\tilde a_{ij}=e^{\boldsymbol L[i,j]-m_i}$,
\[
\Delta\boldsymbol V_i
=\frac{\tilde{\boldsymbol\Phi}^+_i\odot\sum_{j\le i}\tilde a_{ij}\boldsymbol E[i,j]\tilde{\boldsymbol V}_j-\tilde S_i\boldsymbol V_i}{e^{-\log\gamma^2-m_i}+\tilde S_i},
\qquad
\tilde S_i=\sum_{j\le i}\tilde a_{ij}\boldsymbol E[i,j]^2,
\]
which requires one matrix multiplication and two row reductions. Every
exponent is at most zero. With the keys' maximum alone, the null term
$e^{-m_i}\gamma^{-2}$ would overflow when no key agrees with the query, which
is exactly when the null slot should dominate.

\paragraph{Aggregation and update.}
Since $\hat{\boldsymbol A}$ is real, it acts identically on
the real and imaginary halves of the rotated values, as one matrix product on the stacked values,
$\tilde{\boldsymbol M}=\tilde{\boldsymbol V}\hat{\boldsymbol A}^\top$.
The inverse rotation by $\tilde{\boldsymbol\Phi}^+$ gives
\[
\boldsymbol M_{\mathrm{re}}=\boldsymbol C_\omega\odot\tilde{\boldsymbol M}_{\mathrm{re}}-\boldsymbol S_\omega\odot\tilde{\boldsymbol M}_{\mathrm{im}},
\qquad
\boldsymbol M_{\mathrm{im}}=\boldsymbol C_\omega\odot\tilde{\boldsymbol M}_{\mathrm{im}}+\boldsymbol S_\omega\odot\tilde{\boldsymbol M}_{\mathrm{re}} .
\]
The information is a real row sum,
$\tilde S_i=\sum_j\hat{\boldsymbol A}[i,j]\,\boldsymbol E[i,j]$, and the
proximal update divides each query's column by a real scalar,
$\Delta\boldsymbol V_i=\big(\boldsymbol M_i-\tilde S_i\boldsymbol V_i\big)/\big(e^{-m_i}\gamma^{-2}+\tilde S_i\big)$,
acting identically on both halves; $\boldsymbol M_i/\tilde S_i$ is the
estimate $\bar{\boldsymbol V}_i$.
Finally,
$\Delta\boldsymbol Z=\mathcal L_o^{2d\to d}(\Delta\boldsymbol V)$.

\paragraph{Parameterization.}
Each head learns $\sigma^2$, $\eta^2$, $\nu_s$, $\tau$ and $c_h$, kept positive
by a softplus or an exponential; $\mu_h$ and the frequencies are fixed by the
spectrum. We learn the forcing $\sigma^2$ rather than $v_\infty=\sigma^2/(2\mu)$,
so that one formula covers every head,
\[
v_\infty\big(1-\boldsymbol E^2\big)=\sigma^2\,\frac{1-e^{-2\mu\Delta t}}{2\mu},
\]
which tends to $\sigma^2\Delta t$ as $\mu\to0$ and is computed with
$\mathrm{expm1}$ for small $\mu\Delta t$. To decouple the strength of the prior from the absolute scale of the noise, the prior precision is parameterized as $\gamma^{-2}=c_h/\eta^2$. This makes $c_h$ a dimensionless count of agreeing tokens: a key at lag zero in full agreement with the query contributes evidence $1/\eta^2$, so it moves the estimate by $1/(1+c_h)$ whatever the head's decay and noise, and $n$ such keys move it by $n/(n+c_h)$. In Proposition~\ref{prop:window},
$\theta=1/\big(c_h(1-e^{-\mu})\big)$.

\paragraph{Fused implementation.}
Each head performs the two $N\times N$ matrix products of standard attention,
for $\boldsymbol G$ and $\tilde{\boldsymbol M}$; everything else is
elementwise on $N\times N$ matrices or a reduction over rows. The computation
therefore fits a tiled kernel in the style of FlashAttention, which we
implement in Triton. One pass over the keys keeps, for each query, a running
maximum of the logits and two running sums, $\sum_j p_{ij}\tilde{\boldsymbol V}_j$
and $\sum_j p_{ij}\boldsymbol E[i,j]$, with
$p_{ij}=e^{\boldsymbol L[i,j]+\log\boldsymbol E[i,j]-m_i}$, in place of
softmax's value sum and normalizer. The null slot, the division, and the
inverse rotation are applied once the final maximum is known, and the
backward pass recomputes the logits tile by tile.

The overhead comes from the elementwise work on each pair. With the
lag-dependent $\boldsymbol\Sigma_n$ and the Student-$t$ weight, each pair
needs a division and two logarithms, and the forward pass takes about 1.8
times as long as softmax attention in the same kernel. With flat precision
and the exponential weight, the logit adds only a few multiply-adds to the
dot product, and the overhead falls to 10--20\%. With this variant, a training step for a 153M-parameter model at 4096 tokens takes about 1.4
times as long as with RoPE under cuDNN's fused attention, and 1.2 times as
long as with ALiBi; part of the gap is our backward pass, which is slower
than cuDNN's even for softmax attention. The 19.4M-parameter experiments use
the explicit implementation.

\begin{figure}[H]
\centering
\resizebox{0.8\textwidth}{!}{%
\begin{tikzpicture}[
  font=\small, >={Stealth[length=1.8mm]},
  blk/.style={draw, rounded corners=1.5pt, align=center, minimum width=3.6cm,
              minimum height=0.95cm, inner sep=2pt, line width=0.55pt},
  prj/.style={draw=gray!50, rounded corners=1.5pt, align=center, minimum width=3.6cm,
              minimum height=0.55cm, inner sep=1pt, line width=0.45pt,
              font=\scriptsize, text=black!70},
  std/.style={blk, fill=gray!8, draw=gray!55},
  rfa/.style={blk, fill=cyan!10, draw=cyan!45!black},
  dka/.style={blk, fill=orange!18, draw=orange!75!black},
  arr/.style={->, line width=0.6pt, draw=black!70},
  side/.style={->, line width=0.6pt, draw=orange!75!black},
  ttl/.style={font=\normalsize\bfseries},
]
\def\ya{0}\def\yb{1.2}\def\yc{2.6}\def\yd{4.05}\def\ye{5.85}\def\yf{7.45}
\begin{scope}[xshift=0cm]
\node[ttl] at (0,8.35) {RoPE attention};
\node[prj] (a1) at (0,\ya) {Input projections $\boldsymbol W_q,\boldsymbol W_k,\boldsymbol W_v$};
\node[std] (a2) at (0,\yb) {Rotate $\boldsymbol Q,\boldsymbol K$};
\node[std] (a3) at (0,\yc) {Dot product\\$\mathrm{Re}\big(\tilde{\boldsymbol Q}_i^\dagger\tilde{\boldsymbol K}_j\big)$};
\node[std] (a4) at (0,\yd) {Softmax};
\node[std, minimum height=1.75cm] (a5) at (0,\ye) {Average of values};
\node[prj] (a6) at (0,\yf) {Output projection $\boldsymbol W_o$};
\foreach \p/\q in {a1/a2,a2/a3,a3/a4,a4/a5,a5/a6} \draw[arr] (\p) -- (\q);
\draw[arr] (0,-0.75) -- (a1); \draw[arr] (a6) -- (0,8.0);
\end{scope}
\begin{scope}[xshift=4.4cm]
\node[ttl] at (0,8.35) {RFA};
\node[prj] (b1) at (0,\ya) {Input projections $\boldsymbol W_q,\boldsymbol W_k,\boldsymbol W_v$};
\node[rfa] (b2) at (0,\yb) {Rotate $\boldsymbol Q,\boldsymbol K,\boldsymbol V$};
\node[rfa] (b3) at (0,\yc) {Robust residual\\$\|\tilde{\boldsymbol Q}_i-\boldsymbol E\,\tilde{\boldsymbol K}_j\|^2/\boldsymbol\Sigma$};
\node[rfa] (b4) at (0,\yd) {Softmax\\bias $-\log\boldsymbol\Sigma$ (levels off)};
\node[rfa, minimum height=1.75cm] (b5) at (0,\ye) {Average of values,\\counter-rotate $\boldsymbol V$};
\node[prj] (b6) at (0,\yf) {Output projection $\boldsymbol W_o$};
\foreach \p/\q in {b1/b2,b2/b3,b3/b4,b4/b5,b5/b6} \draw[arr] (\p) -- (\q);
\draw[arr] (0,-0.75) -- (b1); \draw[arr] (b6) -- (0,8.0);
\end{scope}
\begin{scope}[xshift=8.8cm]
\node[ttl] at (0,8.35) {DKA};
\node[prj] (c1) at (0,\ya) {Input projections $\boldsymbol W_q,\boldsymbol W_k,\boldsymbol W_v$};
\node[rfa] (c2) at (0,\yb) {Rotate $\boldsymbol Q,\boldsymbol K,\boldsymbol V$};
\node[dka] (c3) at (0,\yc) {Robust residual\\$\|\tilde{\boldsymbol K}_j-\boldsymbol E\,\tilde{\boldsymbol Q}_i\|^2/\boldsymbol\Sigma_n$};
\node[dka] (c4) at (0,\yd) {Softmax + null slot\\bias $\log h=-2\mu\Delta t$};
\node[dka, minimum height=1.75cm] (c5) at (0,\ye) {Average of $\tilde{\boldsymbol V}_j/\boldsymbol E$,\\counter-rotate $\boldsymbol V$;\\[1pt]
  step $\frac{S_i}{\gamma^{-2}+S_i}\big(\bar{\boldsymbol V}_i-\boldsymbol V_i\big)$};
\node[prj] (c6) at (0,\yf) {Output projection $\boldsymbol W_o$};
\foreach \p/\q in {c1/c2,c2/c3,c3/c4,c4/c5,c5/c6} \draw[arr] (\p) -- (\q);
\draw[arr] (0,-0.75) -- (c1); \draw[arr] (c6) -- (0,8.0);
\draw[side] (c1.east) -- ++(0.4,0) |- node[pos=0.27, right, font=\scriptsize,
  text=orange!60!black, align=left] {$\boldsymbol V_i$\\current\\estimate} (c5.east);
\end{scope}
\begin{scope}[shift={(0.9,-1.35)}, font=\scriptsize]
\node[std, minimum width=0.45cm, minimum height=0.3cm] at (0,0) {};
\node[anchor=west] at (0.3,0) {standard attention};
\node[rfa, minimum width=0.45cm, minimum height=0.3cm] at (3.3,0) {};
\node[anchor=west] at (3.6,0) {from RFA's dynamical model};
\node[dka, minimum width=0.45cm, minimum height=0.3cm] at (8.0,0) {};
\node[anchor=west] at (8.3,0) {new in DKA};
\end{scope}
\end{tikzpicture}}
\caption{\textbf{RoPE attention, RFA and DKA.} Gray: standard attention with RoPE;
cyan: introduced by RFA's dynamical model; orange: new in DKA. DKA compares
each key with the query carried back to the key's time, adds ALiBi's bias
from the information each key carries, averages the values rescaled by
$1/\boldsymbol E$ into estimates of the current state, and adds a null slot,
so its output is a step from the current estimate toward the context's
estimate, sized by the evidence.}
\label{fig:dka_overview}
\end{figure}

\newpage

\subsection{Algorithm}
\label{sec:algorithm}

This section gives the full pseudocode for a single isotropic DKA head (Algorithm~\ref{alg:dka}) and a block diagram of it (Figure~\ref{fig:dka_block}).

\begin{algorithm}[H]
\caption{Dynamical Kernel Attention (isotropic, single head)}
\label{alg:dka}
\begin{algorithmic}[1]
\REQUIRE Sequence $\boldsymbol Z\in\mathbb R^{d\times N}$ at positions $t_1,\dots,t_N$
\REQUIRE Linear layers $\mathcal L_q,\mathcal L_k,\mathcal L_v$ ($d\to 2d$) and $\mathcal L_o$ ($2d\to d$)
\REQUIRE Learned scalars $\sigma',\eta',\nu_s,\log\tau,\log c_h$; decay rate $\mu$; frequencies $\omega_1,\dots,\omega_{d/2}$
\ENSURE Updated sequence $\boldsymbol Z^{+}\in\mathbb R^{d\times N}$
\STATE $\boldsymbol\omega \gets (\omega_1,-\omega_1,\dots,\omega_{d/2},-\omega_{d/2})$ \COMMENT{conjugate symmetry}
\STATE $(\sigma^2,\eta^2) \gets \mathrm{softplus}(\sigma',\eta'),\quad \gamma^{-2}\gets c_h/\eta^2$ \COMMENT{variances, prior}
\STATE $\boldsymbol Q,\ \boldsymbol K,\ \boldsymbol V \gets \mathcal L_q(\boldsymbol Z),\ \mathcal L_k(\boldsymbol Z),\ \mathcal L_v(\boldsymbol Z)$ \COMMENT{project}
\STATE $\boldsymbol E[i,j] \gets e^{-\mu|t_i-t_j|},\quad \tilde{\boldsymbol\Phi}^{\pm}[k,i] \gets e^{\pm i\omega_k t_i}$ \COMMENT{decay and rotation}
\STATE $\boldsymbol\Sigma_n \gets \sigma^2(1-\boldsymbol E^{2})/(2\mu)+\eta^2$ \COMMENT{$\to\sigma^2|t_i-t_j|+\eta^2$ as $\mu\to0$}
\STATE $\tilde{\boldsymbol Q},\ \tilde{\boldsymbol K},\ \tilde{\boldsymbol V} \gets \tilde{\boldsymbol\Phi}^{-}\odot\boldsymbol Q,\ \tilde{\boldsymbol\Phi}^{-}\odot\boldsymbol K,\ \tilde{\boldsymbol\Phi}^{-}\odot\boldsymbol V$ \COMMENT{rotate}
\STATE $\|\boldsymbol R_{qk}[i,j]\|^2 \gets \|\boldsymbol K_j\|^2+\boldsymbol E[i,j]^2\,\|\boldsymbol Q_i\|^2-2\,\boldsymbol E[i,j]\,\mathrm{Re}\big(\tilde{\boldsymbol Q}_i^{\dagger}\tilde{\boldsymbol K}_j\big)$ \COMMENT{key-frame residual}
\STATE $\boldsymbol L \gets -\log\boldsymbol\Sigma_n-\tau(\nu_s+1)\log\!\big(1+\tfrac{1}{\nu_s d}\,\|\boldsymbol R_{qk}\|^2\oslash\boldsymbol\Sigma_n\big)+\boldsymbol M_{\le}$ \COMMENT{robust logits}
\STATE $m_i \gets \max\big(\max_j \boldsymbol L[i,j],\,-\log\gamma^2\big),\quad \boldsymbol A[i,j] \gets e^{\boldsymbol L[i,j]-m_i}$ \COMMENT{stabilized weights}
\STATE $\hat{\boldsymbol A} \gets \boldsymbol A\odot\boldsymbol E,\quad \tilde S_i \gets \sum_{j}\hat{\boldsymbol A}[i,j]\,\boldsymbol E[i,j]$ \COMMENT{decay and information}
\STATE $\boldsymbol M \gets \tilde{\boldsymbol\Phi}^{+}\odot\big(\tilde{\boldsymbol V}\hat{\boldsymbol A}^{\top}\big)$ \COMMENT{aggregate, counter-rotate}
\STATE $\Delta\boldsymbol V_i \gets \big(\boldsymbol M_i-\tilde S_i\boldsymbol V_i\big)\big/\big(e^{-\log\gamma^2-m_i}+\tilde S_i\big)$ \COMMENT{proximal innovation}
\STATE $\boldsymbol Z^{+} \gets \boldsymbol Z+\mathcal L_o(\Delta\boldsymbol V)$ \COMMENT{residual update}
\STATE \textbf{return} $\boldsymbol Z^{+}$
\end{algorithmic}
\end{algorithm}

\vspace{0.5cm}

\begin{figure}[H]
  \centering
  \resizebox{\textwidth}{!}{
\begin{tikzpicture}[
  font=\small, >={Stealth[length=2mm]},
  box/.style={draw, rounded corners=2pt, align=center, minimum height=1.05cm, inner sep=3pt, line width=0.6pt},
  std/.style={box, fill=gray!10, draw=gray!60},
  rfa/.style={box, fill=orange!16, draw=orange!75!black},
  dyn/.style={box, fill=blue!7, draw=blue!55!black},
  arr/.style={->, line width=0.7pt, draw=black!75},
  darr/.style={->, line width=0.7pt, draw=blue!55!black},
  skip/.style={->, dashed, line width=0.6pt, draw=black!45},
  lab/.style={font=\scriptsize, inner sep=1pt},
]
\node[std, minimum height=0.8cm] (Z) at (0,-1.35) {$\boldsymbol Z$};
\node[std] (qk)    at (1.8,0)    {$\boldsymbol W_q,\boldsymbol W_k$\\{\scriptsize $d\!\to\!2d$}};
\node[std] (rotqk) at (3.85,0)   {rotate\\$\tilde{\boldsymbol\Phi}^{-}$};
\node[rfa] (res)   at (6.65,0)   {key-frame residual\\$\|\tilde{\boldsymbol K}_j-\boldsymbol E\,\tilde{\boldsymbol Q}_i\|^2$};
\node[rfa] (logit) at (10.35,0)  {robust logits\\$\boldsymbol L=-\log\boldsymbol\Sigma_n\,+$\\$\tau\log w\big(\|\boldsymbol R_{qk}\|^2\oslash\boldsymbol\Sigma_n\big)$};
\node[rfa] (sm)    at (14.1,0)  {unnormalized weights\\$\boldsymbol A=\exp(\boldsymbol L-\boldsymbol m)$};
\node[rfa] (A)     at (17.6,0)  {decay \& information\\$\hat{\boldsymbol A}=\boldsymbol A\odot\boldsymbol E$\\$\tilde S_i=\sum_j\hat{\boldsymbol A}_{ij}\boldsymbol E_{ij}$};

\node[std] (v)     at (1.8,-2.85)   {$\boldsymbol W_v$\\{\scriptsize $d\!\to\!2d$}};
\node[rfa] (rotv)  at (3.85,-2.85)  {rotate\\$\tilde{\boldsymbol\Phi}^{-}$};
\node[std] (agg)   at (6.65,-2.85)  {aggregate\\$\tilde{\boldsymbol V}\hat{\boldsymbol A}^{\top}$};
\node[rfa] (crot)  at (9.55,-2.85)  {counter-rotate\\$\tilde{\boldsymbol\Phi}^{+}$};
\node[rfa] (innov) at (13.15,-2.85) {proximal innovation\\[2pt]$\Delta\boldsymbol V=\dfrac{\boldsymbol M-\tilde{\boldsymbol S}\odot\boldsymbol V}{e^{-\log\gamma^2-\boldsymbol m}+\tilde{\boldsymbol S}}$};
\node[std] (wo)    at (16.45,-2.85) {$\boldsymbol W_o$\\{\scriptsize $2d\!\to\!d$}};
\node[std] (out)   at (19.15,-2.85) {residual update\\$\boldsymbol Z^{+}=\boldsymbol Z+\Delta\boldsymbol Z$};

\node[rfa] (prior) at (9.55,-4.35) {proximal prior\\$\boldsymbol x_i\sim\mathcal N(\boldsymbol V_i,\gamma^2),\ \ \gamma^{-2}=c_h/\eta^2$};

\node[dyn] (E)     at (5.9,2.6)   {decay kernel\\$\boldsymbol E=e^{-\mu\Delta t}$\\{\scriptsize $\mu$ from the spectrum}};
\node[dyn] (par)   at (8.8,2.6)   {per-head scalars\\$\sigma^2,\ \eta^2,\ \tau,\ c_h$};
\node[dyn] (P)     at (12.35,2.6) {conditioned measurement\\$h=\boldsymbol E^2$\\$\boldsymbol\Sigma_n=\tfrac{\sigma^2}{2\mu}(1-\boldsymbol E^2)+\eta^2$};

\draw[arr] (Z.north) |- (qk.west);
\draw[arr] (Z.south) |- (v.west);
\draw[arr] (qk) -- (rotqk);
\draw[arr] (rotqk) -- node[lab, above] {$\tilde{\boldsymbol Q},\tilde{\boldsymbol K}$} (res);
\draw[arr] (res) -- (logit);
\draw[arr] (logit) -- (sm);
\draw[arr] (sm) -- (A);
\draw[arr] ([xshift=-6mm]A.south) -- ([xshift=-6mm]A.south |- 0,-1.2) -- node[lab, above, pos=0.5] {$\hat{\boldsymbol A}$} (agg.north |- 0,-1.2) -- (agg.north);
\draw[arr] ([xshift=6mm]A.south) -- node[lab, right, pos=0.4] {$\tilde{\boldsymbol S}$} ([xshift=6mm]A.south |- 0,-1.65) -- ([xshift=9mm]innov.north |- 0,-1.65) -- ([xshift=9mm]innov.north);
\draw[arr] (v) -- (rotv);
\draw[arr] (rotv) -- node[lab, above] {$\tilde{\boldsymbol V}$} (agg);
\draw[arr] (agg) -- (crot);
\draw[arr] (crot) -- node[lab, above] {$\boldsymbol M$} (innov);
\draw[arr] (innov) -- node[lab, above] {$\Delta\boldsymbol V$} (wo);
\draw[arr] (wo) -- node[lab, above] {$\Delta\boldsymbol Z$} (out);
\draw[arr] (out.east) -- ++(0.7,0);

\draw[skip] (v.south) -- (v.south |- prior.west) -- node[lab, above, pos=0.5] {$\boldsymbol V$ (current estimate)} (prior.west);
\draw[skip] (prior.east) -- (innov.south |- prior.east) -- (innov.south);
\draw[arr] (-1.3,-1.35) -- (Z.west);
\fill[black!60] (-0.75,-1.35) circle (1.3pt);
\draw[skip] (-0.75,-1.35) -- (-0.75,-5.2) -- node[lab, above, pos=0.3] {$\boldsymbol Z$ (residual stream)} (out.south |- 0,-5.2) -- (out.south);

\draw[darr] (par) -- (P);
\draw[darr] (P.north |- 0,3.75) -- (P.north);
\draw[darr] ([xshift=3.5mm]E.south) -- node[lab, left] {$\boldsymbol E$} ([xshift=3.5mm]E.south |- res.north);
\draw[darr] (11.2,0 |- P.south) -- node[lab, right] {$\boldsymbol\Sigma_n$} (11.2,0 |- logit.north);
\draw[darr] (par.south) -- node[lab, left] {$\tau$} (par.south |- logit.north);
\draw[darr] (E.north) -- (E.north |- 0,3.75) -- node[lab, above, pos=0.5] {$\boldsymbol E,\ h$} (A.north |- 0,3.75) -- (A.north);

\begin{scope}[shift={(4.6,-5.95)}, font=\scriptsize]
  \node[std, minimum height=0.3cm, minimum width=0.45cm, inner sep=0pt] at (0,0) {};
  \node[anchor=west] at (0.3,0) {as in RoPE attention};
  \node[rfa, minimum height=0.3cm, minimum width=0.45cm, inner sep=0pt] at (3.6,0) {};
  \node[anchor=west] at (3.9,0) {modified by DKA};
  \node[dyn, minimum height=0.3cm, minimum width=0.45cm, inner sep=0pt] at (6.7,0) {};
  \node[anchor=west] at (7.0,0) {dynamical model (depends on lag only)};
\end{scope}
\end{tikzpicture}
}
\caption{\textbf{A single isotropic DKA head.} The dynamical model (blue)
depends only on the lag, the decay rate $\mu$ set by the spectrum, and a few
learned per-head scalars. It supplies the decay kernel $\boldsymbol E$, the
observation coefficient $h=\boldsymbol E^2$, and the conditioned covariance
$\boldsymbol\Sigma_n$. The query is compared in the key's frame against the
backward conditional mean $\boldsymbol E\tilde{\boldsymbol Q}_i$, and the
temperature $\tau$ sharpens the content term of the logits. Weights are not
softmax-normalized: the proximal prior on the current estimate contributes a
null slot with logit $-\log\gamma^2$, with $\gamma^{-2}=c_h/\eta^2$, to the
denominator, so a head abstains when the accumulated information
$\tilde{\boldsymbol S}$ is small. The row maximum $\boldsymbol m$ includes the
null logit. The robust weight $w$ may be the Student-$t$ or exponential kernel,
and the mask is causal (including the diagonal).}
  \label{fig:dka_block}
\end{figure}

\newpage

\section{Connections to Existing Positional Encodings}
\label{app:pos_encodings}

A single isotropic DKA head contains several positional encodings as limits
(Section~\ref{sec:dka}). RoPE is the limit of zero decay and no forcing, as in
RFA \citep{racioppo2026rfa}. With decay, the head has ALiBi's form exactly,
over the values rescaled by $1/\boldsymbol E$, with a bias that is the
information each key carries. Without decay, its lag bias is the logarithmic
bias of KERPLE, and NoPE and softmax plus one are further limits.
Appendix~\ref{app:other_encodings} relates DKA to other ways of extending RoPE
beyond its training length.

\paragraph{Shared limits.}
Two limits reduce the update to softmax attention. First, the full step
($\gamma^2\to\infty$) removes the null slot, so
$\Delta\boldsymbol V_i=\bar{\boldsymbol V}_i-\boldsymbol V_i$; the term
$-\boldsymbol V_i$ passes through $\mathcal L_o$ as a fixed linear map of the
residual stream, and the attention output is $\bar{\boldsymbol V}_i$, a softmax
over $\boldsymbol L+\log h$ applied to the values rescaled by $1/\boldsymbol E$
(Section~\ref{sec:dka}). Second, a large Student-$t$ parameter
($\nu_s\to\infty$), with $\tau=1$, gives the logits
\[
\boldsymbol L[i,j]=-\log\boldsymbol\Sigma_n[i,j]
-\frac{\|\boldsymbol R_{qk}[i,j]\|^2}{\boldsymbol\Sigma_n[i,j]\,d}.
\]

\paragraph{RoPE.}
With zero decay ($\mu=0$), $\boldsymbol E=\boldsymbol 1$ and $h\equiv1$, so
$\bar{\boldsymbol V}_i$ is a normalized average of the values, transported by
the relative rotation, as in RoPER \citep{harik2022roper}, whereas standard
RoPE leaves them unrotated. The residual is
$\|\boldsymbol K_j\|^2+\|\boldsymbol Q_i\|^2-2\,\mathrm{Re}(\tilde{\boldsymbol Q}_i^\dagger\tilde{\boldsymbol K}_j)$,
and $\boldsymbol\Sigma_n=\sigma^2\Delta t+\eta^2$. With $\sigma^2=0$,
$\boldsymbol\Sigma_n=\eta^2$ is constant, the term $\|\boldsymbol Q_i\|^2$ is the
same for every key and cancels in the normalization, and
\[
\boldsymbol L[i,j]=\frac{2}{\eta^2 d}\,\mathrm{Re}\big(\tilde{\boldsymbol Q}_i^\dagger\tilde{\boldsymbol K}_j\big)
-\frac{\|\boldsymbol K_j\|^2}{\eta^2 d}+\mathrm{const}_i ,
\]
RoPE attention with inverse temperature $2/(\eta^2 d)$ and a bias on the key's
norm, which vanishes for normalized keys (as in QK norm).

\paragraph{ALiBi.}
With decay ($\mu>0$), the estimate is
$\bar{\boldsymbol V}_i=\sum_j\pi_{ij}\,\tilde{\boldsymbol\Phi}^+_i\odot\tilde{\boldsymbol V}_j/\boldsymbol E[i,j]$,
with $\pi_{ij}\propto\exp\big(\boldsymbol L[i,j]-2\mu\Delta t_{ij}\big)$. The
term $\log h_{ij}=-2\mu\Delta t_{ij}$ is ALiBi's linear distance bias
\citep{press2022trainshorttestlong}, exactly and at every lag, with slope
$2\mu$. It comes from the information each key carries about the current
state, so its slope is set by the decay rate rather than by the noise levels.
The bias is applied to estimates rather than to values: transport shrinks each
state toward the equilibrium by $\boldsymbol E$, so a value becomes an
unbiased estimate of the current state only after rescaling by
$1/\boldsymbol E$. As weights on the values themselves, the numerator decays
as $e^{-\mu\Delta t}$ and the normalizer as $e^{-2\mu\Delta t}$, so the
weights can sum to more than one, while a distant key's contribution still
decays. Weighting the values as in softmax attention (D14) gives ALiBi's form
on the values, up to the content logits, the null slot and the value
rotation; its estimate is shrunk toward the equilibrium, and it costs about
1\% in perplexity (Appendix~\ref{app:ablations}). The head also departs from
ALiBi in two bounded ways. First, the lag prior
$-\log\boldsymbol\Sigma_n=-\log\big(v_\infty(1-\boldsymbol E^2)+\eta^2\big)$
lies between $-\log\eta^2$ and $-\log(v_\infty+\eta^2)$: it adds a slope
$2\mu v_\infty/\eta^2$ near zero lag and levels off beyond about $1/(2\mu)$
tokens. Second, for unit-norm queries and keys, the residual is
$(1-\boldsymbol E)^2+2\boldsymbol E\big(1-\mathrm{Re}(\tilde{\boldsymbol Q}_i^\dagger\tilde{\boldsymbol K}_j)\big)$,
so the content term carries the weight $2\boldsymbol E/(\boldsymbol\Sigma_n d)$,
which decays exponentially with lag, as in xPos
\citep{sun2022lengthextrapolatabletransformer}, and even a matching key pays
$(1-\boldsymbol E)^2/(\boldsymbol\Sigma_n d)$, because the backward mean
$\boldsymbol E\tilde{\boldsymbol Q}_i$ shrinks toward the equilibrium. The
linear term is what bounds the evidence from distant tokens
(Appendix~\ref{app:counting_evidence}) and gives each head its memory window
(Appendix~\ref{app:memory_window}).

\paragraph{RFA, Brownian diffusion, and KERPLE.}
Under the diffuse prior, $h\equiv1$, so RFA's normalization has no term
$\log h$, and its only lag bias is $-\log\boldsymbol\Sigma$. In heads without
decay, the latent process is Brownian, $\boldsymbol\Sigma=\sigma^2\Delta t+\eta^2$
grows linearly with lag, and the bias $-\log(1+\sigma^2\Delta t/\eta^2)$ is the
logarithmic variant of KERPLE \citep{chi2022kerple}. As $\mu\to0$ with $\sigma^2$ fixed, $v_\infty\to\infty$ and the stationary
prior becomes diffuse, so DKA's zero-decay heads coincide with RFA's: the same
model gives ALiBi's bias in decaying heads and KERPLE's in undamped ones.

\paragraph{Distant keys in RFA.}
\label{app:rfa_sink}
In a damped RFA head, as $\boldsymbol E\to0$, the residual
$\|\tilde{\boldsymbol Q}_i-\boldsymbol E\tilde{\boldsymbol K}_j\|^2$ tends to
$\|\boldsymbol Q_i\|^2$ and the lag variance to its stationary floor, so the
logits of all distant keys tend to the same content-free value. Their combined
share of the softmax weight therefore grows with the context, while the decay
applied after the softmax multiplies their values by nearly zero: distant keys
form an implicit sink, whose size grows with length. Within the training
window, this sink can act as a learned way to abstain; beyond it, it dilutes
the weight on near keys, and evicting it changes the normalization. In DKA, the
bias $\log h=-2\mu\Delta t$ in the normalization falls without bound, so the
share of distant keys vanishes, and abstention goes to the null slot.

\paragraph{NoPE.}
Under the coupling $\mu_h=b\,\omega_{h,\max}$, every frequency of a head is at
most $\mu_h/b$, so as $\mu_h\to0$ the rotation vanishes along with the decay.
With no forcing, the logits are those of the RoPE limit with
$\tilde{\boldsymbol\Phi}^\pm=\boldsymbol 1$, which contain no positional term:
causal attention without positional encoding (NoPE)
\citep{kazemnejad2023impact}. This limit comes from the coupling, which RFA
shares, rather than from the stationary prior.

\paragraph{Softmax plus one.}
Without the full step, the null slot stays in the denominator of
Eq.~\ref{eq:dka_update}. With zero decay, $h\equiv1$ and
$\boldsymbol E=\boldsymbol 1$, so
\[
\Delta\boldsymbol V_i=\sum_{j\le i}\frac{a_{ij}}{\gamma^{-2}+\sum_{j'\le i}a_{ij'}}
\big(\tilde{\boldsymbol\Phi}^+_i\odot\tilde{\boldsymbol V}_j-\boldsymbol V_i\big),
\]
a softmax over the keys' logits and one extra logit $-\log\gamma^2$ whose
innovation is zero. This is ``softmax plus one'' \citep{miller2023offbyone},
applied to innovations: the extra slot pulls the update toward no change,
whereas in the standard output form it pulls the output toward zero (D19,
Appendix~\ref{app:training}). In decaying heads, the null slot keeps this role
and the keys' logits gain the bias $\log h$.

\subsection{Relation to Other Extrapolation Methods}
\label{app:other_encodings}

Position can enter attention by multiplying the content match or by adding a
bias to it \citep{gu2026deconstructing}; DKA does both.

\paragraph{Rotation with decay.}
xPos \citep{sun2022lengthextrapolatabletransformer} and RetNet
\citep{sun2023retentivenetworksuccessortransformer} both combine RoPE's rotation with exponential decay over the lag. In xPos,
the decay multiplies the logit inside a softmax, which moves a distant key's
logit toward zero rather than below a neutral key's; RetNet has no softmax and
multiplies its weights by the decay. DKA's match is also attenuated, but its
decay also enters additively, through $\log h=-2\mu\Delta t$, and its robust
weight caps what any match can contribute.

\paragraph{Additive biases.}
ALiBi's bias \citep{press2022trainshorttestlong} has been generalized to other
kernels of the lag \citep[KERPLE;][]{chi2022kerple}, to learned functions of it
\citep[FIRE;][]{li2024fire}, and to content-dependent decays
\citep{lin2025forgetting}, and GRAPE \citep{zhang2026group} places such biases
and RoPE's rotations in one group-theoretic framework. DKA instead derives its
bias from the model: ALiBi's in decaying heads, and KERPLE's in undamped ones.

\paragraph{Rescaling frequencies and temperature.}
Most practical methods rescale RoPE's frequencies after training, so that
longer contexts map onto the phases seen in training
\citep{chen2023extending,peng2023yarnefficientcontextwindow,ding2024longrope},
usually with a short fine-tune; without one, YaRN still degrades with length
in our experiments. DKA keeps the frequencies: a head
turns at most $1/b$ radians within its horizon, so its weight has decayed
before its phases leave the range seen in training
(Appendix~\ref{app:spectrum}). YaRN and log-length scaling
\citep{chiang2022overcoming,nakanishi2025scalable} also sharpen the softmax as
the context grows, to counter attention spreading over more keys; in a
decaying DKA head, the evidence summed over keys converges, so this dilution
stops growing beyond the head's memory length.

\newpage

\section{Counting Evidence from Distant Tokens}
\label{app:counting_evidence}

The composite likelihood treats measurements as independent given
$\boldsymbol x_i$, although they share forcing. This section compares how
much evidence DKA and RFA assign to the context with the exact amount, and
how this affects their estimates. The proposition below is the formal
statement of Proposition~\ref{prop:evidence}.

\paragraph{Setting.}
The data are generated by the stationary model of Section~\ref{sec:model}
with $\mu>0$. We work with a single isotropic mode at the first layer, with
unit token spacing, so that $\Delta t_{ij}=i-j$, and consider the $N$ most
recent tokens, $i-N<j\le i$. We set the robust weights to one; since they are
at most one, they can only reduce each token's contribution. We write
\[
\bar\mu:=e^{-2\mu},
\qquad
\ell:=\frac{1}{1-\bar\mu}\approx\frac{1}{2\mu},
\qquad
\sigma_F^2(\Delta t):=v_\infty\big(1-\bar\mu^{\Delta t}\big),
\qquad
s:=\sigma_F^2(1),
\]
for the decay of the variance over one step, the head's memory length, the
variance of the forcing accumulated over a lag, and its value over one step.
Token $j$ is a measurement of the current state (Eq.~\ref{eq:factor}),
\[
\hat x_{ij}=h_{ij}x_i+n_{ij},
\qquad
h_{ij}=\bar\mu^{\Delta t_{ij}},
\qquad
n_{ij}\sim\mathcal N\big(0,\,\bar\mu^{\Delta t_{ij}}\big(\sigma_F^2(\Delta t_{ij})+\eta^2\big)\big).
\]
The rotation in the transport cancels and does not enter.

\paragraph{Information.}
We measure evidence by Fisher information: a measurement $hx+n$ with noise
variance $\Sigma$ carries information $h^2/\Sigma$ about $x$, which in a
Gaussian model is exactly what it adds to the posterior precision. The exact
information $J_N$ accounts for the correlations among the $N$ tokens; a
composite likelihood ignores them and adds each token's individual
information. DKA's factors are the exact individual likelihoods, so it
assigns a token at lag $\Delta t$ the information
\[
I(\Delta t):=\frac{\bar\mu^{\Delta t}}{\sigma_F^2(\Delta t)+\eta^2},
\]
its term in $S_i$ (Section~\ref{sec:dka}), with total
$D_N:=\sum_{\Delta t<N}I(\Delta t)$. RFA treats each token as a direct
measurement, $h=1$, and assigns it $1/\Sigma^{\mathrm{RFA}}(\Delta t)$, where
\[
\Sigma^{\mathrm{RFA}}(\Delta t)
=\bar\mu^{\Delta t}\eta^2+\big(1-\bar\mu^{\Delta t}\big)v_\infty
\]
moves from $\eta^2$ at zero lag to $v_\infty$ at long lag. We omit RFA's floor
$\gamma^2$, which accounts for the query standing in for the latent state;
including it adds $\gamma^2$ to $\eta^2$ and $v_\infty$ without changing the
conclusions.

\paragraph{Estimates.}
Each method's attention output is an estimate $\bar x_i$ of the current
state, toward which the residual connection in RFA and the proximal step in
DKA move part of the way. We compare the estimates themselves, through the
fraction of the current state they retain on average, $\mathbb E[\bar x_i\mid x_i]$.
Under RFA's flat prior, each transported token would satisfy
$\mathbb E[\hat x_{ij}\mid x_i]=x_i$, and RFA's estimate would keep all of the
current state. But no stable process produces a flat distribution: the
stationary model is the one consistent with the dynamics both methods assume.
The two agree at lags short compared with $\ell$, where $h_{ij}\approx1$, and
differ beyond it.

\subsection{Evidence and Bias}

\begin{proposition*}[Formal statement of Proposition~\ref{prop:evidence}]
For every $N$:
\begin{enumerate}
\item[(i)] \emph{Exact information.} $\eta^{-2}\le J_N\le\eta^{-2}+\bar\mu/s$.
\item[(ii)] \emph{DKA's information.} $J_N\le D_N\le\ell/\eta^2$, with
equality on the left for $N\le2$.
\item[(iii)] \emph{RFA's information.} Each token is assigned between
$1/\eta^2$ and $1/v_\infty$, tending to $1/v_\infty$ with lag, so the total
lies between $N/\eta^2$ and $N/v_\infty$.
\item[(iv)] \emph{Bias.} The exact maximum-likelihood estimate and DKA's
estimate satisfy $\mathbb E[\bar x_i\mid x_i]=x_i$. RFA's estimate satisfies
$\mathbb E[\bar x_i\mid x_i]=f_Nx_i$, where $Nf_N\to K$ as $N\to\infty$, with
\[
K:=v_\infty\sum_{\Delta t\ge0}\frac{\bar\mu^{\Delta t}}{\Sigma^{\mathrm{RFA}}(\Delta t)}
\quad\text{between $\ell$ and $\big(v_\infty/\eta^2\big)\ell$.}
\]
\end{enumerate}
\end{proposition*}

\begin{figure}[H]
  \centering
  \begin{subfigure}[b]{0.49\linewidth}
    \centering
    \includegraphics[width=\linewidth]{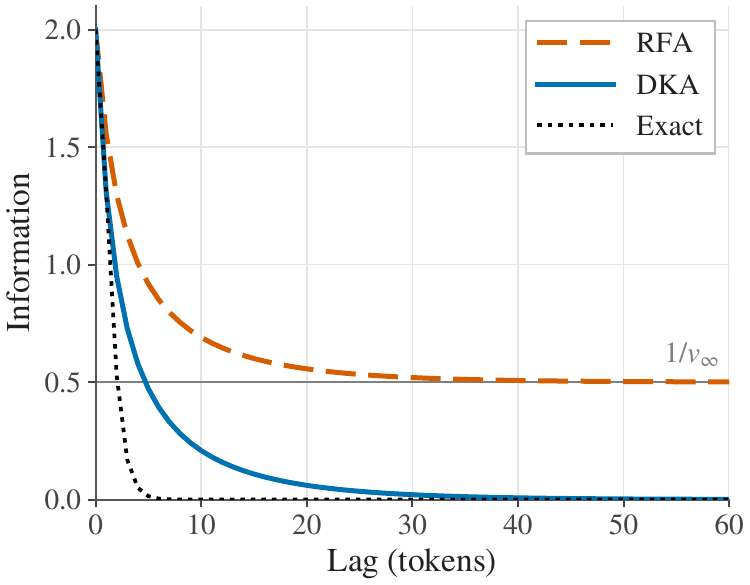}
    \caption{Information per token, by lag}
    \label{fig:counting_evidence_lag}
  \end{subfigure}
  \hfill
  \begin{subfigure}[b]{0.49\linewidth}
    \centering
    \includegraphics[width=\linewidth]{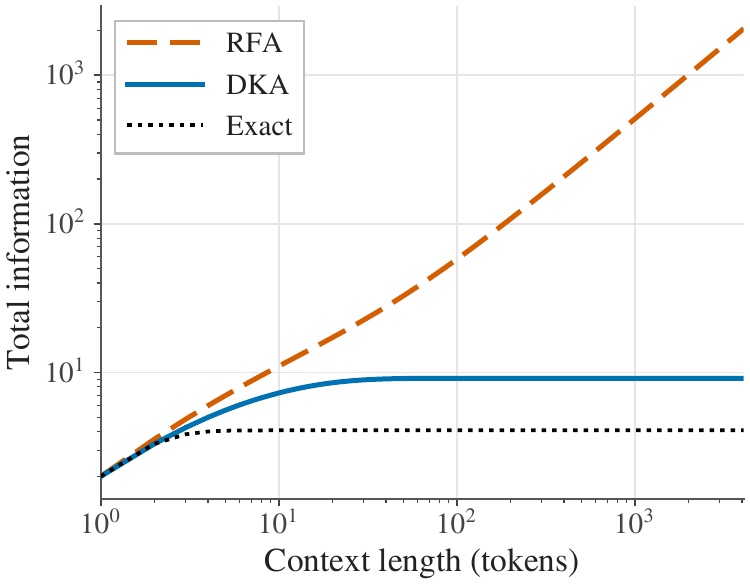}
    \caption{Total information, by context length}
    \label{fig:counting_evidence_total}
  \end{subfigure}
      \caption{\textbf{Counting evidence from distant tokens.} One mode of the
  linear Gaussian model, with stationary data and unit weights $w_{ij}=1$
  ($\mu=0.05$, $v_\infty=2$, $\eta^2=0.5$).
  (a) Information about the current state assigned to one token. DKA's falls
  exponentially with lag; RFA's levels off at $1/v_\infty$, since a distant
  token restates the prior. The exact curve is a token's information given all
  newer tokens.
  (b) Total information in the $N$ most recent tokens. The exact and DKA
  totals saturate, DKA's above the exact by a bounded factor; RFA's grows
  linearly.}
  \label{fig:counting_evidence}
\end{figure}

\paragraph{Interpretation.}
A distant token reports only that $x_i$ is a draw from the stationary
distribution, which the prior already says; RFA counts this once per token,
so its total grows without bound. DKA still overcounts, because nearby tokens
share forcing: all older tokens reach the current state through the same last
step of forcing, so together they give at most $\bar\mu/s$, but DKA applies
this limit to each token separately. Applying it separately always leaves at
least as much, so DKA can overcount but never undercount
(Figure~\ref{fig:counting_evidence}). The counts set the bias. RFA averages
all $N$ tokens, but only about $K$ still carry the current state, and the rest
vote for the equilibrium. DKA weights each token by $h_{ij}$ in both numerator and denominator, so distant tokens drop out. With few tokens, the three estimates agree, RFA's error slightly below the exact one because it trades
bias for variance. As the context grows, each further token adds a vote for
the equilibrium in RFA's average, so its error rises toward that of guessing
the equilibrium, while DKA's levels off above the exact error
(Figure~\ref{fig:counting_evidence_error}).

\paragraph{Proof of (i).}
The current token $\hat x_{ii}=x_i+\varepsilon_i$ has noise independent of
the older tokens, so its information adds to theirs: $J_N=\eta^{-2}+G_N$,
where $G_N$ is the older tokens' information about $x_i$, and
$J_1=\eta^{-2}$.

The older tokens are the $N-1$ most recent tokens at position $i-1$, so by
stationarity they give $x_{i-1}$ posterior precision $v_\infty^{-1}+J_{N-1}$.
Since $x_i=\Phi_{i,i-1}x_{i-1}+F_{i,i-1}$, with $|\Phi_{i,i-1}|^2=\bar\mu$ and
$F_{i,i-1}\sim\mathcal N(0,s)$ independent of the older tokens, the
posterior variance of $x_i$ given them is, writing $J:=J_{N-1}$ and using
$s=v_\infty(1-\bar\mu)$,
\[
\frac{\bar\mu}{v_\infty^{-1}+J}+s
=\frac{\bar\mu v_\infty}{1+v_\infty J}+v_\infty(1-\bar\mu)
=v_\infty\,\frac{\bar\mu+(1-\bar\mu)(1+v_\infty J)}{1+v_\infty J}
=v_\infty\,\frac{1+sJ}{1+v_\infty J}.
\]
Its inverse, the posterior precision, is
\[
\frac{1}{v_\infty}\cdot\frac{1+v_\infty J}{1+sJ}
=\frac{1}{v_\infty}\Big(1+\frac{(v_\infty-s)J}{1+sJ}\Big)
=v_\infty^{-1}+\frac{\bar\mu J}{1+sJ},
\]
since $v_\infty-s=\bar\mu v_\infty$. Information is what the tokens add to
the prior precision $v_\infty^{-1}$, so the older tokens' information about
$x_i$ is the second term:
\[
G_N=\varphi(J_{N-1}),
\qquad
\varphi(u):=\frac{\bar\mu u}{1+su}=\frac{\bar\mu}{u^{-1}+s}.
\]
The bounds follow from $0\le\varphi(u)\le\bar\mu/s$ for all $u\ge0$.

\paragraph{Proof of (ii).}
\emph{Upper bound.} Since $\sigma_F^2\ge0$, the denominator of
$I(\Delta t)$ is at least $\eta^2$, so $I(\Delta t)\le\bar\mu^{\Delta t}/\eta^2$,
and the geometric series gives
\[
D_N=\sum_{\Delta t<N}I(\Delta t)
\le\frac{1}{\eta^2}\sum_{\Delta t\ge0}\bar\mu^{\Delta t}
=\frac{1}{\eta^2}\cdot\frac{1}{1-\bar\mu}
=\frac{\ell}{\eta^2}.
\]

\emph{Lower bound.} The forcing over $\Delta t+1$ steps is the forcing over
$\Delta t$ steps plus one new step, decayed over the other $\Delta t$:
\[
\sigma_F^2(\Delta t)+\bar\mu^{\Delta t}s
=v_\infty\big(1-\bar\mu^{\Delta t}\big)+\bar\mu^{\Delta t}v_\infty(1-\bar\mu)
=v_\infty\big(1-\bar\mu^{\Delta t+1}\big)
=\sigma_F^2(\Delta t+1).
\]
Evaluating $\varphi$ at $u=I(\Delta t)$, whose inverse is
$\bar\mu^{-\Delta t}\big(\sigma_F^2(\Delta t)+\eta^2\big)$, and multiplying
the numerator and denominator by $\bar\mu^{\Delta t}$,
\[
\varphi\big(I(\Delta t)\big)
=\frac{\bar\mu}{\bar\mu^{-\Delta t}\big(\sigma_F^2(\Delta t)+\eta^2\big)+s}
=\frac{\bar\mu^{\Delta t+1}}{\sigma_F^2(\Delta t)+\bar\mu^{\Delta t}s+\eta^2}
=\frac{\bar\mu^{\Delta t+1}}{\sigma_F^2(\Delta t+1)+\eta^2}
=I(\Delta t+1):
\]
carrying a token's information forward one step gives the information of the
token one lag older. Separating the current token, with $I(0)=\eta^{-2}$,
and applying this identity to the older tokens,
\begin{align*}
D_N=\sum_{\Delta t=0}^{N-1}I(\Delta t)
&=\eta^{-2}+\sum_{\Delta t=1}^{N-1}\varphi\big(I(\Delta t-1)\big)
=\eta^{-2}+\sum_{\Delta t=0}^{N-2}\varphi\big(I(\Delta t)\big)\\
&\ge\eta^{-2}+\varphi\Big(\sum_{\Delta t=0}^{N-2}I(\Delta t)\Big)
=\eta^{-2}+\varphi(D_{N-1}).
\end{align*}
The second line uses $\varphi(a)+\varphi(b)\ge\varphi(a+b)$, which holds
because $\varphi$ is concave with $\varphi(0)=0$; the last step is the
definition of $D_{N-1}$.

We now show $D_N\ge J_N$ by induction. For $N=1$, both totals contain only
the current token, so $D_1=J_1=\eta^{-2}$. Suppose $D_{N-1}\ge J_{N-1}$.
Since $\varphi$ is increasing (a larger $u$ makes $u^{-1}$ smaller),
\[
D_N\;\ge\;\eta^{-2}+\varphi(D_{N-1})
\;\ge\;\eta^{-2}+\varphi(J_{N-1})
\;=\;J_N,
\]
where the equality is the recursion from the proof of (i). For $N=2$, the
sum over older tokens has a single term, so the first inequality is an
equality, and $D_2=J_2$.

\paragraph{Proof of (iii).}
$\Sigma^{\mathrm{RFA}}(\Delta t)$ is a weighted average of $\eta^2$ and
$v_\infty$, with the weight on $v_\infty$ rising to one. Each token's
information $1/\Sigma^{\mathrm{RFA}}(\Delta t)$ therefore lies between
$1/\eta^2$ and $1/v_\infty$ and tends to $1/v_\infty$; summing $N$ of them
gives the bounds on the total.

\paragraph{Proof of (iv).}
Each token has mean $\mathbb E[\hat x_{ij}\mid x_i]=h_{ij}x_i$.

\emph{Exact.} Stack the tokens as $\hat{\boldsymbol x}=\boldsymbol h\,x_i+\boldsymbol n$,
where $\boldsymbol n\sim\mathcal N(\boldsymbol 0,\boldsymbol\Sigma^{\mathrm{joint}})$,
so that
$J_N=\boldsymbol h^\dagger\big(\boldsymbol\Sigma^{\mathrm{joint}}\big)^{-1}\boldsymbol h$.
Setting the derivative of the negative log-likelihood,
$\tfrac12(\hat{\boldsymbol x}-\boldsymbol hx)^\dagger\big(\boldsymbol\Sigma^{\mathrm{joint}}\big)^{-1}(\hat{\boldsymbol x}-\boldsymbol hx)$,
to zero gives the maximum-likelihood estimate
\[
\bar x_i=J_N^{-1}\,\boldsymbol h^\dagger\big(\boldsymbol\Sigma^{\mathrm{joint}}\big)^{-1}\hat{\boldsymbol x},
\qquad\text{so}\qquad
\mathbb E[\bar x_i\mid x_i]
=J_N^{-1}\,\boldsymbol h^\dagger\big(\boldsymbol\Sigma^{\mathrm{joint}}\big)^{-1}\boldsymbol h\,x_i
=x_i .
\]

\emph{DKA.} With unit weights and a full step (Eq.~\ref{eq:dka_update}),
DKA's estimate weights each token by its information,
$\bar x_i=D_N^{-1}\sum_jI(\Delta t_{ij})\,\hat x_{ij}/h_{ij}$. Each
$\hat x_{ij}/h_{ij}$ has mean $x_i$, and the weights $I(\Delta t_{ij})/D_N$
sum to one, so $\mathbb E[\bar x_i\mid x_i]=x_i$.

\emph{RFA.} With unit weights, RFA's estimate is a normalized average of the
transported tokens themselves,
\[
\bar x_i=\sum_j\pi_{ij}\hat x_{ij},
\qquad
\pi_{ij}=\frac{1/\Sigma^{\mathrm{RFA}}(\Delta t_{ij})}{\sum_{j'}1/\Sigma^{\mathrm{RFA}}(\Delta t_{ij'})}.
\]
The weights do not depend on the data, so
$\mathbb E[\bar x_i\mid x_i]=\big(\sum_j\pi_{ij}h_{ij}\big)x_i$, and the
bracket is $f_N$. Substituting the weights and $h_{ij}=\bar\mu^{\Delta t_{ij}}$,
\[
f_N=\frac{A_N}{B_N},
\qquad
A_N:=\sum_{\Delta t<N}\frac{\bar\mu^{\Delta t}}{\Sigma^{\mathrm{RFA}}(\Delta t)},
\qquad
B_N:=\sum_{\Delta t<N}\frac{1}{\Sigma^{\mathrm{RFA}}(\Delta t)} .
\]
The numerator $A_N$ counts the weight on tokens that still carry the current
state. Its terms are $\bar\mu^{\Delta t}$ times RFA's per-token information,
which lies between $1/v_\infty$ and $1/\eta^2$ by (iii). So $A_N$ converges,
to $K/v_\infty$ by the definition of $K$, and summing the geometric series
places $K$ between $\ell$ and $(v_\infty/\eta^2)\ell$. The denominator $B_N$
counts the weight on all $N$ tokens. Its terms tend to $1/v_\infty$, so
$B_N/N\to1/v_\infty$. Hence,
\[
Nf_N=\frac{A_N}{B_N/N}\;\longrightarrow\;\frac{K/v_\infty}{1/v_\infty}=K .
\]
\hfill$\square$

\newpage

\section{Memory Window}
\label{app:memory_window}

Because past tokens remain available, forgetting in DKA is a property of a
token's evidence rather than of its storage. This gives a rule for when
storage can be released: each head can discard tokens older than a fixed
window, with a guaranteed bound on the change to its output that holds for
every query.

\paragraph{Setting.}
We consider a single isotropic head with decay $\mu>0$, tokens at integer
positions, and the update of Eq.~\ref{eq:dka_update},
\[
\Delta\boldsymbol V_i=\frac{\boldsymbol U_i}{c+S_i},
\qquad
\boldsymbol U_i:=\sum_{j\le i}a_{ij}\boldsymbol E[i,j]\,\hat{\boldsymbol v}_{ij}-S_i\boldsymbol V_i,
\qquad
S_i:=\sum_{j\le i}a_{ij}\boldsymbol E[i,j]^2,
\]
with $c:=\gamma^{-2}=c_h/\eta^2$,
$\hat{\boldsymbol v}_{ij}:=\tilde{\boldsymbol\Phi}^+_i\odot\tilde{\boldsymbol V}_j$
and $a_{ij}=w_{ij}^\tau/\boldsymbol\Sigma_n[i,j]$. Two facts hold for every
query and every token. First, $0\le w_{ij}\le1$, so
$0\le a_{ij}\le1/\boldsymbol\Sigma_n[i,j]$, whatever the content. Second, the
rotations have unit magnitude, so $\|\hat{\boldsymbol v}_{ij}\|=\|\boldsymbol V_j\|\le B$.
We write $\boldsymbol E(\Delta t)$ and $\boldsymbol\Sigma_n(\Delta t)$ for the
values at lag $\Delta t$; each lag holds at most one key.

\begin{proposition*}[Proposition~\ref{prop:window}, restated]
For every query $i$, every context length, every choice of content weights,
and every $T\ge1$,
\[
\big\|\Delta\boldsymbol V_i-\Delta\boldsymbol V_i^{(T)}\big\|\le B\,B_h(T),
\qquad
B_h(T)=\frac1c\sum_{\Delta t\ge T}
\frac{\boldsymbol E(\Delta t)+(2+K_h)\,\boldsymbol E(\Delta t)^2}{\boldsymbol\Sigma_n(\Delta t)},
\]
where
\(
K_h=\frac1c\sum_{\Delta t\ge0}\frac{\boldsymbol E(\Delta t)}{\boldsymbol\Sigma_n(\Delta t)},
\)
and $B_h(T)\le\theta(3+\theta)\,e^{-\mu T}$ with $\theta=1/\big(c_h(1-e^{-\mu})\big)$.
\end{proposition*}

\paragraph{Proof.}
Let $\boldsymbol U_i'$ and $S_i'$ be the sums over the kept keys, at lag below
$T$, so that $\Delta\boldsymbol V_i^{(T)}=\boldsymbol U_i'/(c+S_i')$.

\emph{What the evicted keys contribute.} Subtracting the truncated sums from
the full ones leaves only the evicted keys:
\[
S_i-S_i'=\sum_{\Delta t_{ij}\ge T}a_{ij}\boldsymbol E[i,j]^2,
\qquad
\boldsymbol U_i-\boldsymbol U_i'
=\sum_{\Delta t_{ij}\ge T}a_{ij}\boldsymbol E[i,j]\big(\hat{\boldsymbol v}_{ij}-\boldsymbol E[i,j]\boldsymbol V_i\big).
\]
Since $\|\hat{\boldsymbol v}_{ij}-\boldsymbol E[i,j]\boldsymbol V_i\|\le B(1+\boldsymbol E[i,j])$
and $a_{ij}\le1/\boldsymbol\Sigma_n[i,j]$,
\[
\big\|\boldsymbol U_i-\boldsymbol U_i'\big\|\le B\sum_{\Delta t\ge T}\frac{\boldsymbol E+\boldsymbol E^2}{\boldsymbol\Sigma_n},
\qquad
0\le S_i-S_i'\le\sum_{\Delta t\ge T}\frac{\boldsymbol E^2}{\boldsymbol\Sigma_n}.
\]

\emph{How the update changes.} Writing the difference of the two ratios over
a common denominator,
\[
\Delta\boldsymbol V_i-\Delta\boldsymbol V_i^{(T)}
=\frac{\boldsymbol U_i-\boldsymbol U_i'}{c+S_i}
-\Delta\boldsymbol V_i^{(T)}\,\frac{S_i-S_i'}{c+S_i}.
\]
Both denominators are at least $c$. The truncated update satisfies
\[
\big\|\Delta\boldsymbol V_i^{(T)}\big\|
\le\frac{B\sum_ja_{ij}\boldsymbol E[i,j]+BS_i'}{c+S_i'}
\le\frac Bc\sum_{\Delta t\ge0}\frac{\boldsymbol E}{\boldsymbol\Sigma_n}+B
=B(1+K_h),
\]
using $c+S_i'\ge c$ in the first term and $S_i'/(c+S_i')\le1$ in the second.
Combining,
\[
\big\|\Delta\boldsymbol V_i-\Delta\boldsymbol V_i^{(T)}\big\|
\le\frac Bc\sum_{\Delta t\ge T}\frac{\boldsymbol E+\boldsymbol E^2}{\boldsymbol\Sigma_n}
+\frac{B(1+K_h)}{c}\sum_{\Delta t\ge T}\frac{\boldsymbol E^2}{\boldsymbol\Sigma_n}
=B\,B_h(T).
\]

\emph{Closed form.} Since $\boldsymbol\Sigma_n\ge\eta^2$, $1/c=\eta^2/c_h$ and
$\boldsymbol E^2\le\boldsymbol E$, each term of $B_h(T)$ is at most
$(3+K_h)\,e^{-\mu\Delta t}/c_h$, and $K_h\le\theta$ by the same argument.
Summing the geometric tail,
$\sum_{\Delta t\ge T}e^{-\mu\Delta t}=e^{-\mu T}/(1-e^{-\mu})$, gives
$B_h(T)\le\theta(3+\theta)e^{-\mu T}$.
\hfill$\square$

\begin{figure}[H]
  \centering
  \begin{subfigure}[b]{0.49\linewidth}
    \centering
    \includegraphics[width=\linewidth]{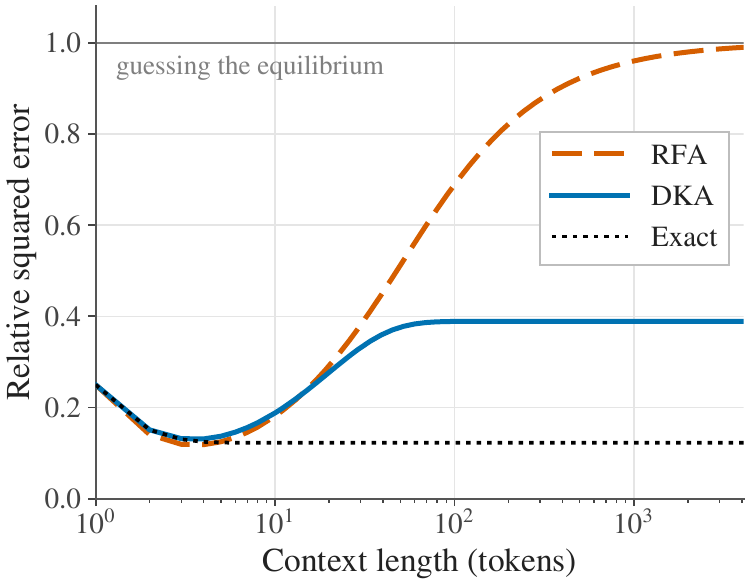}
    \caption{Error of the estimate, by context length}
    \label{fig:counting_evidence_error}
  \end{subfigure}
  \hfill
  \begin{subfigure}[b]{0.49\linewidth}
    \centering
    \includegraphics[width=\linewidth]{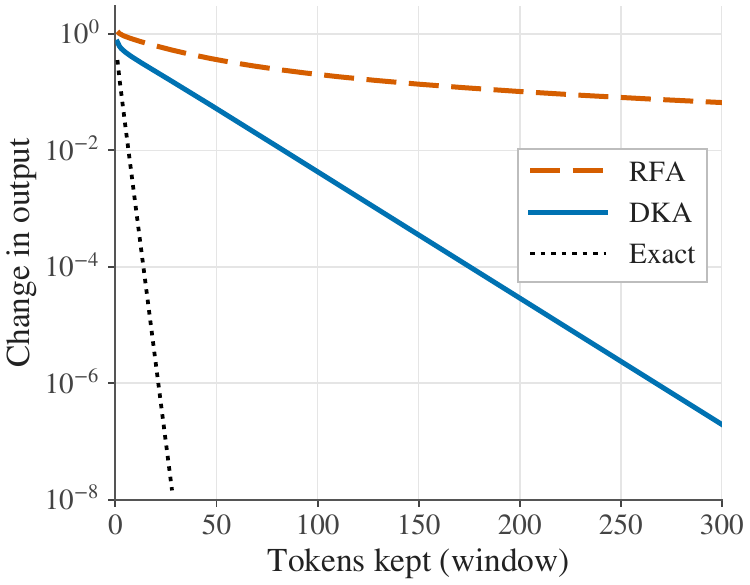}
    \caption{Change in output from evicting old tokens}
    \label{fig:counting_evidence_eviction}
  \end{subfigure}
    \caption{\textbf{Accuracy and memory} (setting as in
Figure~\ref{fig:counting_evidence}), for the estimates $\bar x_i$ before the
  residual or proximal step.
    (a) Mean squared error $\mathbb E\big[(\bar x_i-x_i)^2\big]/v_\infty$; one
  is the error of always guessing the equilibrium.
  (b) RMS change in the estimate, relative to $\sqrt{v_\infty}$, when only the $T$ most recent tokens of a $2048$-token context are kept.}
  \label{fig:counting_evidence_accuracy}
\end{figure}

\paragraph{Scope of the bound.}
The bound uses only $w_{ij}^\tau\le1$, so it holds for any content weights;
content can suppress a key sooner, but only for some queries. The cap
$a_{ij}\le1/\boldsymbol\Sigma_n$ limits what each evicted key contributes, and
the null slot keeps the denominator at least $c$, so the kept keys are not
amplified; as $c_h\to0$, $B_h(T)$ grows without bound. The bound applies to
each layer's attention output, not to the final loss, and is conservative for
slow heads, where $K_h$ is large, and tight for fast ones. In the setting of
Proposition~\ref{prop:evidence}, the change from eviction falls exponentially
for DKA, so a fixed window suffices for every query; it falls slowly for RFA,
whose normalizer does not decay a distant token's weight, and fastest for the
exact estimate (Figure~\ref{fig:counting_evidence_eviction}). A bound that
holds for every query is equivalent to a limit on retrieval: no match can make
an evicted key matter. 

\newpage

\section{Experimental Details}
\label{app:training}

\paragraph{Model.}
All models are 6-layer pre-norm decoders with width 256, 8 heads, a
feed-forward width of 1024, and tied input and output embeddings, using the
GPT-2 tokenizer (vocabulary 50{,}257). Each model has 19.4M parameters, of
which 12.9M are in the embedding. As in RFA \citep{racioppo2026rfa}, every
model, including the baselines, lifts each head to a complex space of
dimension 32 (16 conjugate frequency pairs) through $d\to2d$ projections and
maps back through a $2d\to d$ output projection
(Appendix~\ref{app:implementation}), so parameter counts match exactly. DKA
adds five scalars per head ($\sigma^2$, $\eta^2$, $\nu_s$, $\tau$, $c_h$).
DKA and RFA use the spectrum $\mu_h=b\,\omega_{h,\max}$ with $b=1$ and RoPE's
base $10^4$; each head takes 16 consecutive frequencies of the 128.

\paragraph{Training.}
\begin{table}[h]
\centering
\caption{Training configuration of the 19.4M parameter models, shared by all of them.}
\label{tab:training}
\small
\begin{tabular}{ll}
\toprule
Dataset & WikiText-103 \\
Training length & 512 tokens \\
Epochs / tokens & 15 / 1.76B \\
Batch size & 16 \\
Optimizer & Adam, $\beta=(0.9,0.999)$ \\
Peak learning rate & $10^{-3}$ \\
Schedule & 5\% linear warm-up, cosine decay \\
Gradient clipping & 1.0 \\
\bottomrule
\end{tabular}
\end{table}
DKA's per-head scalars are initialized to a stationary variance
$\sigma^2/2\mu_h=0.5$ in every head (so $\sigma^2=\mu_h$), $\eta^2=1$,
$\nu\approx4d+2$, $\tau=1$ and $c_h=1$, so that the null slot starts at the
weight of one agreeing token.

\paragraph{Evaluation.}
Test perplexity is computed on non-overlapping blocks of 512, 1024, 2048, 4096
and 8192 tokens (547, 271, 134, 64 and 30 blocks), with no fine-tuning and
no position rescaling. At long lengths, attention is computed over blocks of
queries, which gives the same result with less memory. For YaRN
\citep{peng2023yarnefficientcontextwindow}, the position rescaling is applied
at inference without the short fine-tune it normally uses. Paired comparisons
use the per-block mean loss of two models on the same blocks and report the
mean percentage change in perplexity with its standard error over blocks.

\paragraph{Cache eviction.}
Eviction is evaluated on positions 2048--4096 of 32 test blocks of 4096
tokens. Calibrated windows are set per head from 8 validation blocks, as the
smallest lag beyond which the head's measured weight stays below
$\varepsilon=0.01$. Bound windows use $T_h(\varepsilon)=\min\{T:B_h(T)\le\varepsilon\}$
with the exact bound of Proposition~\ref{prop:window}, at $\varepsilon=0.01$.
Capped windows are limited to 512 keys. ALiBi's slope windows keep, in each
head, the lags below $\ln(1/\varepsilon)/m_h$, where $m_h$ is the head's slope,
so that the bias of a dropped key is at most $\ln\varepsilon$ (this ignores
how the content logits vary with lag). The worst-case error is the 99.9th
percentile, over tokens, of the increase in per-token loss relative to the
same model with full context.

\paragraph{Ablations.}

D1 returns the weighted average $M_i/S_i$ in place of the innovation; D2
skips the counter-rotation of the values.
D3 replaces the Student-$t$
weight by its light-tailed limit, $\exp(-d_{ij}^2/d)$. D4 keeps D3's weight but
scores keys by the dot product: it drops the norm terms $\|\tilde K_j\|^2$ and
$E_{ij}^2\|\tilde Q_i\|^2$ of the residual and keeps only the cross term
$-2E_{ij}\,\mathrm{Re}\langle\tilde Q_i,\tilde K_j\rangle$, with everything else as in DKA. D5 also drops the
attenuation of the match, so the content logit is $2\,\mathrm{Re}\langle\tilde Q_i,\tilde K_j\rangle/(\nu\Sigma_n)$ at
every lag; the values are DKA's, each key's estimate $U_{ij}V_j/E_{ij}$. D6 fixes
$\Sigma_n=\eta^2$ at every lag, so the bias $-\log\Sigma_n$ is constant and
the residual is scaled the same at every lag, while the decay still enters
through $E$. D7 combines D6 with the exponential weight of D3, so the logit is
$\log E_{ij}-\tau\|\tilde K_j-E_{ij}\tilde Q_i\|^2/(\nu\eta^2)$: the cheapest
form for a fused kernel.
D8 sets $\gamma^{-2}=0$, so the update reduces to a normalized
average. D9 gives every head all of RoPE's frequencies and keeps its decay;
D10 also applies p-RoPE \citep{barbero2025round}, setting to zero the
frequencies whose period exceeds the training length in heads whose horizon
exceeds it. D11 sets every $\omega=0$, keeping the decays. D12 sets
$\mu_h=0$ in every head and keeps each head's forcing
$\sigma_h^2=2\mu_h v_\infty$ from its coupled decay, so
$\Sigma_n=\eta^2+\sigma^2\Delta t$ grows linearly with lag and the precision
falls as a power law. D13 learns each head's decay, starting from the coupled
value $\mu_h=\omega_{h,\max}$: heads whose frequencies all complete a cycle
within the training length may decay down to a horizon of $L$, and the others
may only decay faster, since slower decay helps within the window at a cost
beyond it that training cannot see.

D14--D17 go from DKA to rotary attention with ALiBi slopes, one assumption of
the model at a time. DKA weights each key's estimate of the current state,
$U_{ij}V_j/E_{ij}$, by $a_{ij}h_{ij}$, so the weights on the values,
$a_{ij}E_{ij}/(\gamma^{-2}+S_i)$, can sum to more than one. D14 uses the
value $U_{ij}V_j$ itself, with the same weight $a_{ij}h_{ij}$, so the weights
on the values sum to at most one, as in softmax attention. D15 also scores
keys by the dot product without the attenuation of the match (D5's score), with the
exponential weight: the content logit is
$2\,\mathrm{Re}\langle\tilde Q_i,\tilde K_j\rangle/(\nu\Sigma_n)$. D16
removes the rest of the noise model: the bias $-\log\Sigma_n$, the scaling of
the content by $1/\Sigma_n$ and its learned inverse temperature (replaced by
$1/\sqrt d$), and the counter-rotation of the values. What remains is rotary
attention with the pre-softmax bias $\log h_{ij}=-2\mu_h\Delta t_{ij}$, in
DKA's innovation form with a null slot holding $V_i$, as in D19. D17 removes
the null slot and the innovation, returning the weighted average: rotary
attention with the bias $-2\mu_h\Delta t$, which is ALiBi with slopes derived
from the spectrum. At zero decay, with all of RoPE's frequencies in every head, the
same reductions give RoPE attention, which DKA's implementation reproduces
exactly (Appendix~\ref{app:pos_encodings}).

\paragraph{Baselines with DKA's components.}
D18 subtracts the current value from RoPE's attention output before the output
projection, $W_o(\sum_j w_{ij}V_j-V_i)$, as DKA's update does. D19 adds DKA's
null slot: an extra logit $b_h$, learned per head from 0, whose value is
$V_i$, so the output is $W_o\sum_j w_{ij}(V_j-V_i)$ with
$\sum_j w_{ij}=1-w_{i,\mathrm{null}}$. As in DKA, the null slot pulls the
update toward no change; a null logit in the standard output form
\citep{miller2023offbyone} would instead pull the output toward zero. D20 gives
D18 DKA's spectrally coupled frequencies, each head a band of RoPE's, with no
decay, and D21 adds the bias $-2\mu_h\Delta t$ its decay implies (D17 in
innovation form). B3 is RFA with its published estimate output; D22 uses the
innovation output, and D23 adds DKA's null slot to it.

\newpage

\section{Additional Experimental Results}
\label{app:additional_experiments}

\subsection{All Ablations}
\label{app:ablations}

Table~\ref{tab:ablation} lists every ablation; Appendix~\ref{app:training}
defines each one.

\paragraph{Null slot and bounded scores.}

Without the null slot (D8), the drift comes from the slowest head: the
normalization must place weight on distant tokens when no nearby token
agrees, and limiting that head alone to lags below 512 brings perplexity at
4096 tokens back to 26.53. With the null slot, the evidence's share of the
update, $S_i/(\gamma^{-2}+S_i)$, ranges from 0.03 to 0.17 across heads, so the
update adds evidence rather than renormalizing it.
D17's drift comes from its middle heads. Dropping every head's keys at lag
512 and beyond, at 4096 tokens, lowers the loss on positions 512--4096 by 0.19
nats per token; the heads with horizons 316 and 1000 account for 0.12 and 0.07
of it (the two of one layer alone for 0.10), and the slowest head, with
horizon 3162, for none. At lag 512, their decay bias $-2\mu_h\Delta t$ is only
$-3.2$ and $-1.0$, too weak to offset the dot-product scores of keys at lags
never seen in training, which nothing bounds; with the same spectrum, DKA's
distant keys help (Table~\ref{tab:probe}).

With the values combined as in softmax attention (D14), scoring keys by the
dot product (D15) gives the best perplexity of any DKA variant within the
window (27.52) but breaks beyond it (54.81 at 2048 tokens, 167.16 at 8192), as
D4 drifts with DKA's values: without the floor of the residual, nothing bounds
a distant match. With DKA's values and no attenuation (D5), the same score
diverges ($7.3\cdot10^7$ at 4096 tokens), since each value's weight can reach
$1/E$. Adding the null slot to D17 (D16) barely helps within the
window (27.76 against 27.86 at 512 tokens).

\paragraph{Lean variant.}

The lean variant used at scale (D7) costs about as much as the exponential
weight alone within the window (27.83 against 27.82 for D3 and 27.59 for
DKA), and the costs of the two simplifications add beyond it (25.92 at 8192
tokens, against 25.67 for D3 and 25.42 for D6). It keeps the extrapolation,
improving with context to 8192 tokens, and the cache result (806 keys per
head, worst-case error 0.13, against 862 and 0.12).

\paragraph{Output and estimator.}

The dot-product score (D4) helps within the window (27.57 at 512 tokens,
against 27.82 for D3, whose weight it shares) before drifting beyond 2048
tokens. The output form matters within the window: averaging the values
instead of taking the innovation (D1) costs $0.2$ at 512 tokens and $0.4$ at
4096. Removing the value rotation (D2) changes little: $+0.2\pm0.2\%$ at 512
tokens, and 26.00 at 4096 against 25.64. In the estimator, replacing the
Student-$t$ weight by the exponential (D3) costs $0.2$ at 512 tokens and $0.4$
at 4096. The Student-$t$ weight works in its heavy-tailed regime even though
the learned $\nu/d$ is 6 to 23: weighted by their share of the update, keys
have $u=R^2/(\nu\Sigma_n)$ between 0.26 and 1.24 (10th to 90th percentile,
median 0.50), where its log-weight $-\kappa\log(1+u)$ is 0.65 to 0.89 of its
Gaussian limit $-\kappa u$.

\paragraph{Forcing, decay, and rotation.}

The forcing is small but used: fixing $\Sigma_n=\eta^2$ (D6) costs about $0.17$
at every length (27.76 at 512 tokens and 25.42 at 8192, against 27.59 and
25.25), although it changes a key's precision by at most 10\%
(Appendix~\ref{app:interpretability}). The decay is what makes DKA
extrapolate: with zero decay and the forcing kept (D12), so that $\Sigma_n$
grows linearly with lag, DKA matches its full model within the window (27.77)
but fails beyond it (118.09 at 2048 tokens). Without rotations (D11), DKA
loses $0.3$ within the window (27.88 at 512 tokens) and drifts slightly beyond
it (26.93 at 8192, against 25.25): the decay alone keeps it nearly stable, and
the rotations improve it at every length. Learning the decay (D13) changes
nothing: 27.59 at 512 tokens and 25.23 at 8192, with the same cache (866 keys
per head, worst-case error 0.09). The three fastest heads, free to slow down to
a horizon of 512 tokens, settle at $\mu_h=b\,\omega_{h,\max}$ with $b=0.4$--$0.9$,
close to the coupling. Started ten times weaker ($b=0.1$), they strengthen only
to $b\approx0.2$, and the model is $0.45\pm0.18\%$ worse within the window: the
optimum is flat, but it favors decay near the coupled value.

\paragraph{DKA's components in the baselines.}

Adding DKA's components to the baselines (D18--D23) shows which of them
transfer. DKA's innovation output helps RoPE within the window: RoPE in
innovation form (D18) is the best model we trained at 512 tokens (27.43,
against 27.59 for DKA; $-0.46\pm0.21\%$ paired over 547 blocks), but it fails
beyond the training length (1834 at 2048 tokens). It also nearly removes RFA's
drift (D22: 27.36 at 8192 tokens, against 30.02 for B3), and DKA's null slot
adds nothing further (D23: 27.99). Nor does it help RoPE: D19 matches D18 within
the window (27.45) and fails sooner beyond it (4100 at 2048 tokens).

\paragraph{Spectral coupling.}

Whether coupling the spectrum helps depends on the decay. Without decay, the
coupled bands cost RoPE 1.0\% at 512 tokens (D20: 27.70, against 27.43 for
D18): a head confined to one band cannot place its attention as freely in lag.
With DKA's decay, the uncoupled spectrum is what costs (D9: 28.15, 2.0\% worse
than DKA), and it fails beyond the window: each band's decay keeps its head
within the lags its frequencies resolve. The decay bias the bands imply (D21)
costs RoPE a further 0.6\% (27.88) and only partly stabilizes it beyond the
window (39.53 at 8192 tokens, against 33.45 for D17 in the estimate form).
DKA, with the same bands and decays, is better than both at every length
(27.59 at 512, 25.25 at 8192).


\begin{table}[H]
\centering
\caption{All ablations at $b=1$. Test perplexity at 512 to 8192 tokens (trained at 512), and the cache eviction result (keys
per head and worst-case error, the 99.9th percentile increase in per-token loss, with calibrated windows). D1--D13 each change one part of DKA; D14--D17 are
cumulative, each including the changes above it (D16 keeps the decay bias and the null slot, D17 only the bias); D18--D23 add DKA's components to the baselines B1--B3, listed
for reference.}
\label{tab:ablation}
\small
\setlength{\tabcolsep}{5.75pt}
\begin{tabular}{llccccccc}
\toprule
Model & Change from DKA & 512 & 1024 & 2048 & 4096 & 8192 & Keys & Worst \\
\midrule
DKA   & --                                   & 27.59 & 26.39 & 25.75 & 25.64 & 25.25 & 862  & 0.12 \\
\midrule
D1    & estimate output                      & 27.80 & 26.54 & 25.90 & 26.07 & 26.16 & 887  & 0.09 \\
D2    & no value rotation                    & 27.65 & 26.53 & 26.02 & 26.00 & 25.64 & 764  & 0.23 \\
\midrule
D3    & exponential weight                   & 27.82 & 26.67 & 26.10 & 26.05 & 25.67 & 798  & 0.11 \\
D4    & dot-product score                    & 27.57 & 26.37 & 26.26 & 28.95 & 36.05 & 739  & 2.32 \\
D5    & dot product, unattenuated & 27.51 & 26.53 & 56564 & $7.3\cdot10^{7}$ & $2.9\cdot10^{9}$ & --   & -- \\
D6    & flat precision                       & 27.76 & 26.56 & 25.91 & 25.82 & 25.42 & 851  & 0.08 \\
D7    & flat precision + exp.\ weight & 27.83 & 26.81 & 26.33 & 26.28 & 25.92 & 806  & 0.13 \\
D8    & no null slot                         & 27.54 & 26.38 & 26.76 & 32.36 & 63.36 & 523  & 1.72 \\
\midrule
D9    & not coupled                          & 28.15 & 27.92 & 41.88 & 195.23 & 1762 & 1102 & 0.35 \\
D10   & not coupled + p-RoPE                 & 28.13 & 27.23 & 28.68 & 38.44 & 61.41 & 1024 & 0.12 \\
D11   & no rotations                         & 27.88 & 26.75 & 26.39 & 26.79 & 26.93 & 697  & 0.19 \\
D12   & zero decay                           & 27.77 & 27.90 & 118.09 & 2753 & 17979 & 2451 & 0.50 \\
D13   & learned decay                        & 27.59 & 26.38 & 25.73 & 25.63 & 25.23 & 866  & 0.09 \\
\midrule
D14   & values as in softmax                 & 27.89 & 26.70 & 26.03 & 25.90 & 25.49 & 710  & 0.03 \\
D15   & + dot-product score                  & 27.52 & 26.48 & 54.81 & 115.10 & 167.16 & 957  & 0.44 \\
D16   & + no noise model & 27.76 & 27.04 & 37.60 & 50.97 & 74.27 & 923  & 2.23 \\
D17   & + no null slot & 27.86 & 26.92 & 28.04 & 31.02 & 33.45 & 928  & 2.71 \\
\midrule
B1    & RoPE                                 & 27.93 & 40.55 & 100.63 & 202.36 & 305.69 & --   & -- \\
B2    & ALiBi                                & 28.63 & 27.33 & 26.57 & 26.41 & 25.99 & 321  & 1.28 \\
B3    & RFA                                  & 27.74 & 26.47 & 26.01 & 27.15 & 30.02 & 509  & 10.92 \\
\midrule
D18   & RoPE + innovation                    & 27.43 & 188.19 & 1834 & 9775 & 22325 & --   & -- \\
D19   & D18 + null slot                      & 27.45 & 374.56 & 4100 & 20199 & 44022 & --   & -- \\
D20   & D18, coupled frequencies & 27.70 & 160.60 & 1475 & 17492 & 259551 & --   & -- \\
D21   & D20 + decay bias                     & 27.88 & 26.78 & 30.45 & 34.21 & 39.53 & 895  & 2.65 \\
D22   & RFA + innovation                     & 27.78 & 26.65 & 26.41 & 26.87 & 27.36 & 234  & 4.54 \\
D23   & D22 + null slot                      & 28.03 & 26.88 & 26.39 & 26.79 & 27.99 & 190  & 3.48 \\
\bottomrule
\end{tabular}
\end{table}

\subsection{Cache Eviction}
\label{app:eviction}

\begin{table}[h]
\centering
\caption{Cache eviction at 4096 tokens: keys kept per head (of 3073 on average with full context), perplexity on
positions 2048--4096, and worst-case error (99.9th percentile increase in per-token loss against the model's own
full context). Capped: each head's window limited to 512 keys, the training length. Slope windows: per-head
windows set from ALiBi's slopes.}
\label{tab:eviction}
\small
\begin{tabular}{llccc}
\toprule
Model & Policy & Keys & Perplexity & Worst-case error \\
\midrule
RoPE   & sliding window       & 512  & 21.74 & 5.38 \\
ALiBi  & full context         & --   & 22.03 & -- \\
ALiBi  & sliding window       & 512  & 22.35 & 4.04 \\
ALiBi  & calibrated           & 321  & 22.13 & 1.28 \\
ALiBi  & slope windows        & 294  & 22.21 & 1.83 \\
RFA    & full context         & --   & 23.81 & -- \\
RFA    & sliding window       & 512  & 21.62 & 4.18 \\
RFA    & calibrated           & 509  & 41.88 & 10.92 \\
\midrule
DKA    & full context         & --   & 21.43 & -- \\
DKA    & sliding window       & 512  & 21.49 & 2.92 \\
DKA    & calibrated           & 862  & 21.43 & 0.12 \\
DKA    & bound                & 1382 & 21.43 & 0.004 \\
DKA    & bound, capped        & 324  & 21.49 & 2.92 \\
\bottomrule
\end{tabular}
\end{table}

Table~\ref{tab:eviction} lists the eviction results of
Section~\ref{sec:exp_eviction}. The bound windows grow as
$\ln(1/\varepsilon)/\mu_h$, so $\varepsilon$ barely trades cache for quality
(every $\varepsilon\le1$ is lossless): the bound mainly gives each head a
window in proportion to its horizon, and it is conservative for slow heads and tight for fast ones, and applies to each layer's attention output, not to
the final loss. In the setting of Proposition~\ref{prop:evidence}
(Figure~\ref{fig:counting_evidence_eviction}), the change from eviction falls
exponentially for DKA, at a rate set by $\mu$, so a fixed window suffices for
every query. For RFA it falls slowly, since a distant token's weight in its
normalizer is not decayed. The exact estimate's change falls fastest: once
recent states are pinned down, older tokens add almost nothing.

\begin{table}[h]
\centering
\caption{Per-head windows against a sliding window that keeps the same number of keys per head, on the same DKA
checkpoint: perplexity on positions 2048--4096 at 4096 tokens and worst-case error (Table~\ref{tab:eviction}).
DKA's full context: 21.43.}
\label{tab:eviction_matched}
\small
\begin{tabular}{lccccc}
\toprule
& & \multicolumn{2}{c}{Per-head windows} & \multicolumn{2}{c}{Sliding window} \\
\cmidrule(lr){3-4} \cmidrule(lr){5-6}
Per-head policy & Keys & Perplexity & Worst-case & Perplexity & Worst-case \\
\midrule
calibrated, capped ($\varepsilon=0.01$) & 281  & 21.50 & 2.93  & 21.98 & 5.35 \\
bound, capped ($\varepsilon=0.01$)      & 324  & 21.49 & 2.92  & 21.83 & 4.83 \\
calibrated ($\varepsilon=0.03$)         & 646  & 21.45 & 0.53  & 21.42 & 1.82 \\
calibrated ($\varepsilon=0.01$)         & 862  & 21.43 & 0.12  & 21.37 & 0.86 \\
bound ($\varepsilon=0.01$)              & 1382 & 21.43 & 0.004 & 21.38 & 0.29 \\
\bottomrule
\end{tabular}
\end{table}

Table~\ref{tab:eviction_matched} fixes the model and the cache size and
changes only how the keys are chosen. With small caches, per-head windows
capped at the training length are better than a sliding window: they keep few
keys in the fast heads and up to 512 in the slow ones, and they reach the
perplexity of a 512-key sliding window (21.50 and 21.49 against 21.49) with
45\% and 37\% fewer keys.

From 646 keys on, a
sliding window matches DKA's full context or slightly improves on it (21.37
at 862 keys), so at 4096 tokens, the slow heads' keys beyond about 650 tokens
slightly hurt. The per-head windows reproduce the full-context model instead,
and their advantage at these sizes is the worst-case error: 0.12 against 0.86
at 862 keys, and 0.004 against 0.29 at 1382. With a 512-key sliding window,
DKA's worst-case error is 2.92, against 5.38 for RoPE, 4.04 for ALiBi and 4.18
for RFA.

ALiBi also gets per-head windows, either calibrated by the same rule as DKA's
or set from its slopes (Appendix~\ref{app:training}), so the comparison with
DKA is between per-head policies. ALiBi's slope windows keep 294 keys for
22.21 and 441 keys for 22.06 (worst-case errors 1.83 and 0.57), and its
calibrated windows keep 321 keys for 22.13 and 502 for 22.04 (1.28 and 0.21).
At every cache size, DKA's perplexity is lower: 21.50 with 281 keys and 21.49
with 324, below ALiBi's 22.03 with its full context of 3073 keys.

\subsection{Perplexity by Position}
\label{app:positions}

Perplexity averaged over a 512-token block mixes positions with very different
amounts of context. Table~\ref{tab:position} splits the paired comparison by
target position, over all 547 test blocks of 512 tokens. DKA is better than
every baseline on the first 64 positions, by 3--6\%. Beyond them, it still beats
RoPE and ALiBi, and ties RFA. The null slot changes nothing within the window.

\begin{table}[h]
\centering
\caption{Change in test perplexity relative to DKA (\%, positive is worse than DKA) by target position within
512-token blocks, mean $\pm$ standard error over 547 WikiText-103 test blocks.}
\label{tab:position}
\small
\begin{tabular}{lcccc}
\toprule
Model & 0--64 & 64--256 & 256--512 & All \\
\midrule
B1, RoPE   & $+3.1\pm0.6$ & $+0.6\pm0.4$ & $+1.2\pm0.3$ & $+1.2\pm0.2$ \\
B2, ALiBi  & $+5.7\pm0.6$ & $+2.9\pm0.3$ & $+3.9\pm0.3$ & $+3.7\pm0.2$ \\
B3, RFA    & $+3.6\pm0.6$ & $+0.3\pm0.3$ & $+0.0\pm0.3$ & $+0.5\pm0.2$ \\
\midrule
D2         & $+0.1\pm0.5$ & $-0.2\pm0.3$ & $+0.6\pm0.2$ & $+0.2\pm0.2$ \\
D8         & $+0.2\pm0.6$ & $-0.3\pm0.3$ & $-0.1\pm0.3$ & $-0.2\pm0.2$ \\
D17        & $+3.0\pm0.6$ & $+0.8\pm0.4$ & $+0.6\pm0.3$ & $+1.0\pm0.2$ \\
\bottomrule
\end{tabular}
\end{table}

\subsection{Choice of Spectrum}
\label{app:spectrum}

Table~\ref{tab:spectrum} varies $b$ at RoPE's base, so the frequencies stay
fixed and only the decay changes.  Faster
decay ($b=2$, $b=5$) extrapolates but is worse at every length, as faster heads are able to see less of the context. Slower decay
($b=0.5$) fails beyond 1024 tokens.

\begin{table}[h]
\centering
\caption{Spectra at RoPE's base $10^4$ (horizons $1/\mu_h$ in tokens, slowest first): test perplexity.}
\label{tab:spectrum}
\small
\begin{tabular}{rlccccc}
\toprule
$b$ & Horizons & 512 & 1024 & 2048 & 4096 & 8192 \\
\midrule
0.5 & 6325, 2000, 632, 200, 63, 20, 6, 2       & 27.66 & 26.54 & 28.29 & 31.84 & 35.10 \\
1   & 3162, 1000, 316, 100, 32, 10, 3, 1       & \textbf{27.59} & \textbf{26.39} & \textbf{25.75} & \textbf{25.64} & \textbf{25.25} \\
2   & 1581, 500, 158, 50, 16, 5, 1.6, 0.5      & 27.92 & 26.84 & 26.30 & 26.23 & 25.85 \\
5   & 632, 200, 63, 20, 6.3, 2, 0.6, 0.2       & 27.96 & 26.84 & 26.25 & 26.14 & 25.74 \\
\bottomrule
\end{tabular}
\end{table}

To find where the far context helps or hurts, we drop each head's keys at lags
beyond 512 at 4096 tokens, one head (or one head index in every layer) at a
time, and measure the change in loss on positions 512--4096
(Table~\ref{tab:probe}). At $b=1$, all the gain from distant context comes
from the two slowest heads, mostly in one layer; the other six are fully
decayed by lag 512. At $b=0.5$, the whole failure comes from one band, the
head with horizon 632, just above the training length: dropping its distant
keys alone brings perplexity on positions 512--4096 from 29.67 to about 22.8,
the level of $b=1$, while the two slowest heads still help. That head's
fastest frequency has a period of about 2000 tokens: within the window its
phase turns 1.6 radians, and beyond it, the phase reaches angles never seen in
training while its attenuation $E=e^{-\Delta t/632}$ is still 0.44 at lag
512. At $b=1$, the same
frequencies decay twice as fast, so the head has little weight left where its
phases are new. This is the constraint the coupling enforces: a head turns at
most $1/b$ radians within its horizon. One might expect the two slowest heads
(horizons 6325 and 2000) to cause the failure, since their evidence grows the
most beyond the training length (up to $1/(1-e^{-2\mu_hL})=6.7$ times); they
help instead.

\begin{table}[h]
\centering
\caption{Change in loss (milli-nats per token, positions 512--4096 at 4096 tokens, 32 test blocks) when one head
index, in every layer, drops its keys at lag $\ge512$. Positive: the distant keys helped. Heads are ordered from
slowest to fastest.}
\label{tab:probe}
\small
\begin{tabular}{lrrrrrr}
\toprule
Model & All & Head 0 & Head 1 & Head 2 & Head 3 & Heads 4--7 \\
\midrule
$b=0.5$ & $-264\pm24$ & $+19.6$ & $+14.5$ & $-261.9$ & $-15.5$ & $0.0$ \\
$b=1$   & $+5.3\pm1.7$ & $+4.9$ & $+2.1$ & $-0.1$ & $0.0$ & $0.0$ \\
$b=2$   & $-0.4\pm0.5$ & $-1.8$ & $+2.4$ & $+0.1$ & $0.0$ & $0.0$ \\
\bottomrule
\end{tabular}
\end{table}

\subsection{Attention Maps}
\label{app:attn_maps}

\begin{figure}[h]
\centering
\begin{subfigure}[b]{\linewidth}
  \centering
  \includegraphics[width=\linewidth]{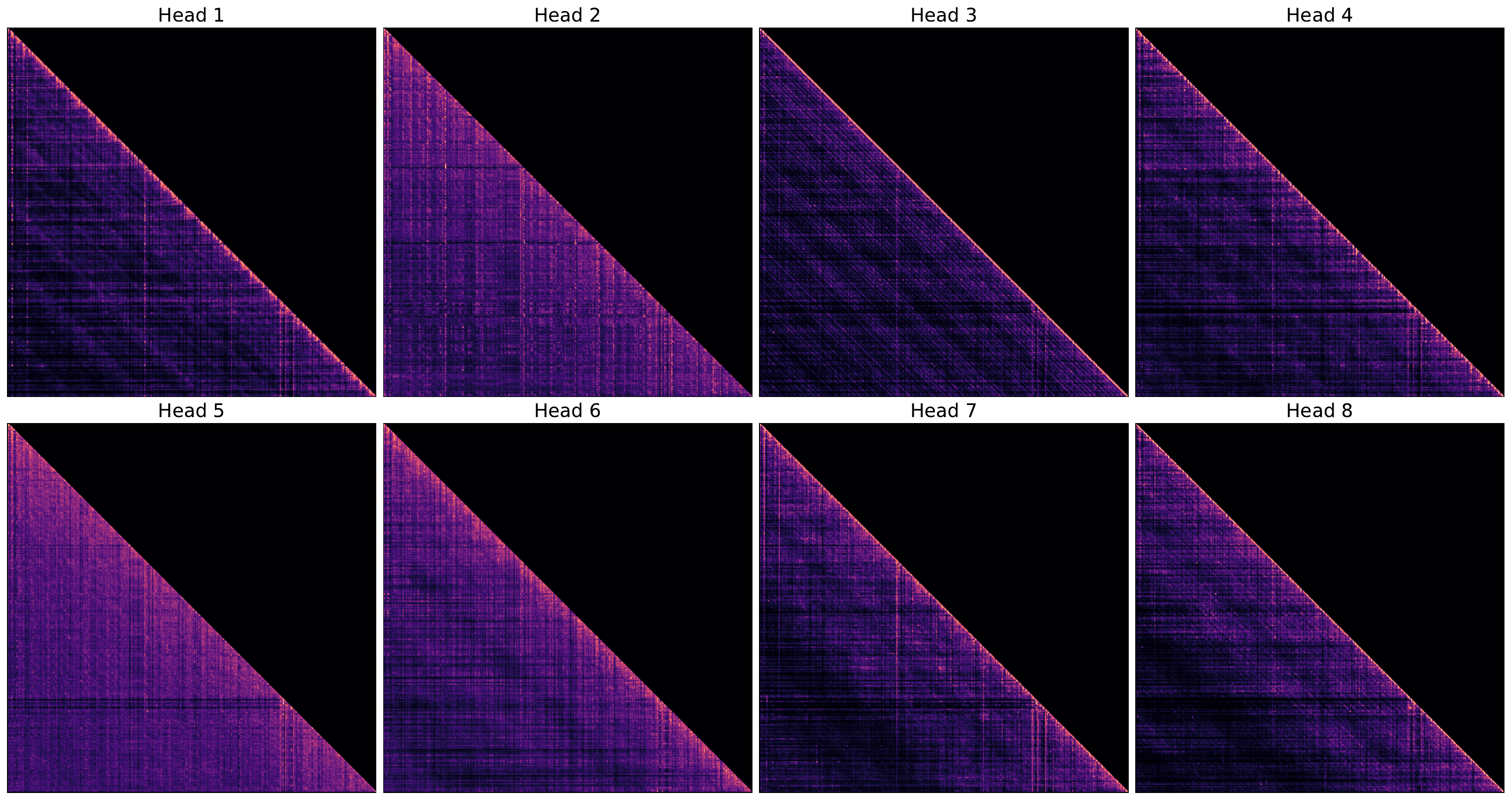}
  \caption{RoPE (B1): softmax weights of each head.}
  \label{fig:attn_maps_rope}
\end{subfigure}

\vspace{0.6em}
\begin{subfigure}[b]{\linewidth}
  \centering
  \includegraphics[width=\linewidth]{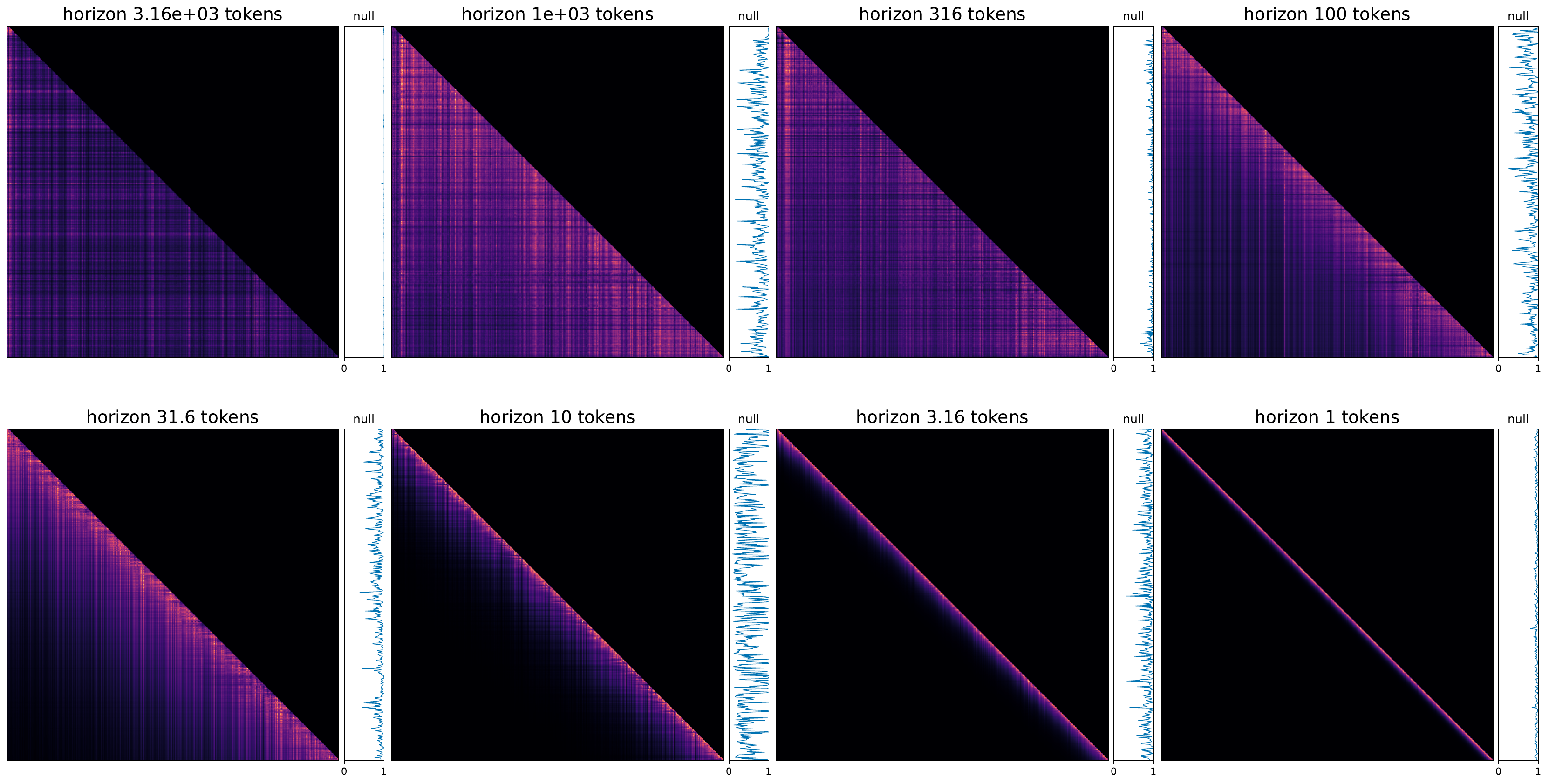}
  \caption{DKA: weight of each value in the update, $a_{ij}E_{ij}/(\gamma^{-2}+S_i)$, heads ordered by horizon;
  beside each map, the null slot's share $\gamma^{-2}/(\gamma^{-2}+S_i)$ for each query.}
  \label{fig:attn_maps_dka}
\end{subfigure}
\caption{Attention maps of the last block on the first 512-token block of the WikiText-103 test split (within the
training length), color scale $x^{0.2}$.}
\label{fig:attn_maps}
\end{figure}

Figure~\ref{fig:attn_maps} compares the last block's attention maps of RoPE and
DKA within the training length. Every RoPE head spreads its weight over the
whole window, with diagonal bands from its rotations. DKA's heads form a bank
of scales, from the slowest, which weights the whole window, to the fastest,
which attends only to the previous few tokens, and none shows periodic bands:
with $b=1$, each head turns at most one radian within its horizon, so its
weight decays before a period of its frequencies can show.

\subsection{Learned Parameters and Evidence}
\label{app:interpretability}

\begin{figure}[h]
\centering
\begin{subfigure}[b]{0.32\linewidth}
  \includegraphics[width=\linewidth]{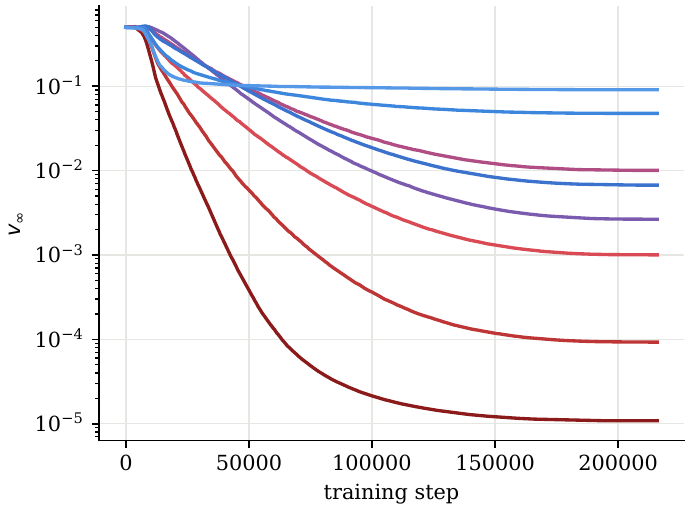}
  \caption{$v_\infty=\sigma^2/2\mu_h$}
  \label{fig:dka_params_vinf}
\end{subfigure}\hfill
\begin{subfigure}[b]{0.32\linewidth}
  \includegraphics[width=\linewidth]{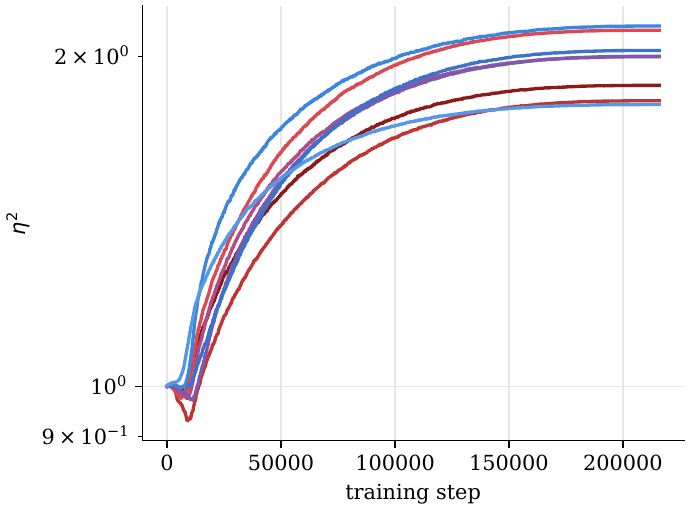}
  \caption{$\eta^2$}
  \label{fig:dka_params_eta}
\end{subfigure}\hfill
\begin{subfigure}[b]{0.32\linewidth}
  \includegraphics[width=\linewidth]{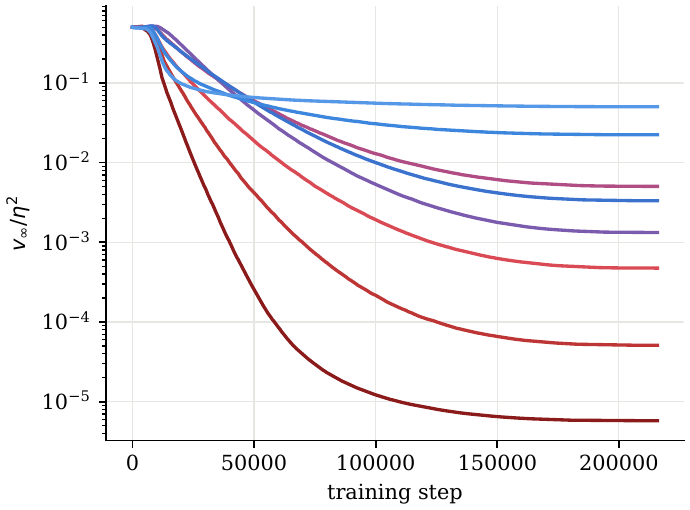}
  \caption{$v_\infty/\eta^2$}
  \label{fig:dka_params_snr}
\end{subfigure}

\begin{subfigure}[b]{0.32\linewidth}
  \includegraphics[width=\linewidth]{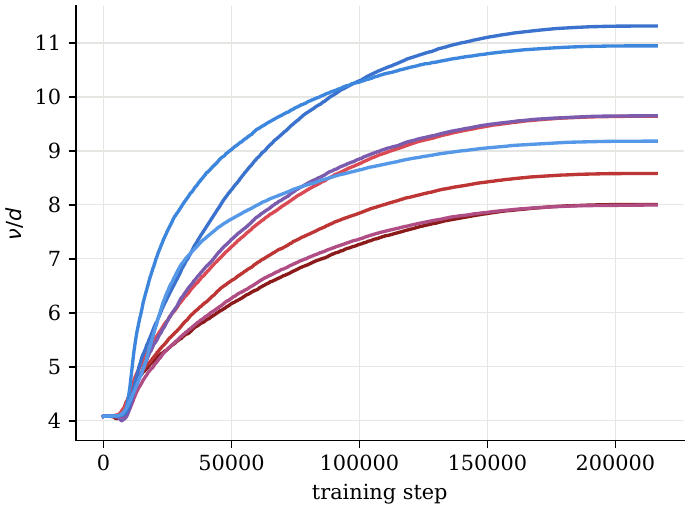}
  \caption{$\nu/d$}
  \label{fig:dka_params_nu}
\end{subfigure}\hfill
\begin{subfigure}[b]{0.32\linewidth}
  \includegraphics[width=\linewidth]{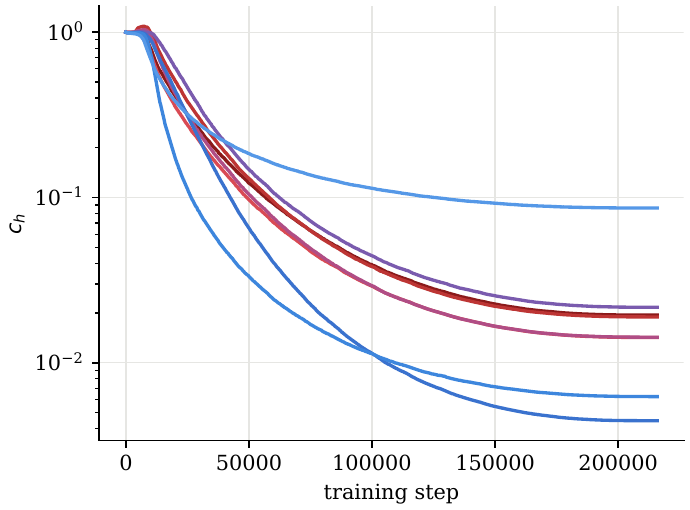}
  \caption{$c_h$}
  \label{fig:dka_params_c}
\end{subfigure}\hfill
\begin{subfigure}[b]{0.32\linewidth}
  \includegraphics[width=\linewidth]{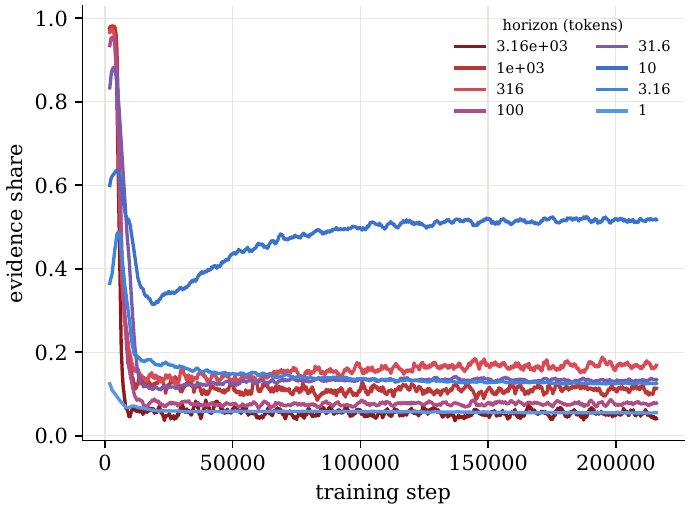}
  \caption{evidence share $S_i/(\gamma^{-2}+S_i)$}
  \label{fig:dka_params_share}
\end{subfigure}
\caption{DKA's learned parameters over training in the last of the six blocks, one line per head, from the
slowest (dark red) to the fastest (blue). The evidence share (f) is a per-batch mean, smoothed over 2000 steps.}
\label{fig:dka_params}
\end{figure}

Figure~\ref{fig:dka_params} shows how DKA's parameters evolve during training.
The forcing becomes small: in the last block, $v_\infty$ falls by more than
four orders of magnitude in the slowest head while $\eta^2$ doubles, and
$v_\infty/\eta^2$ ends between
$10^{-5}$ in the slowest head and $0.05$ in the fastest, so a key's precision
falls with lag by at most 5\% (10\% in the first block), and recency comes
mostly from the attenuation $E^2$ that the decay imposes. The forcing still
helps: fixing $\Sigma_n=\eta^2$ (D6) costs about $0.17$ at every length. In
the same block, the Student-$t$ parameter $\nu/d$ rises from 4 to 8--11, the
null slot's weight $c_h$ falls from one agreeing token to between 0.005 and
0.09, and the heads' initial differences in how much they abstain disappear
within the first 10k steps.

\begin{figure}[h]
\centering
\begin{subfigure}[b]{0.49\linewidth}
  \includegraphics[width=\linewidth]{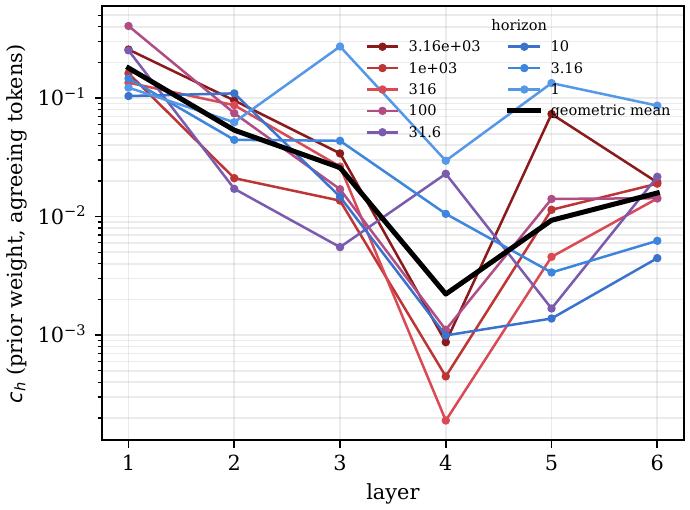}
  \caption{$c_h$}
  \label{fig:dka_depth_c}
\end{subfigure}\hfill
\begin{subfigure}[b]{0.49\linewidth}
  \includegraphics[width=\linewidth]{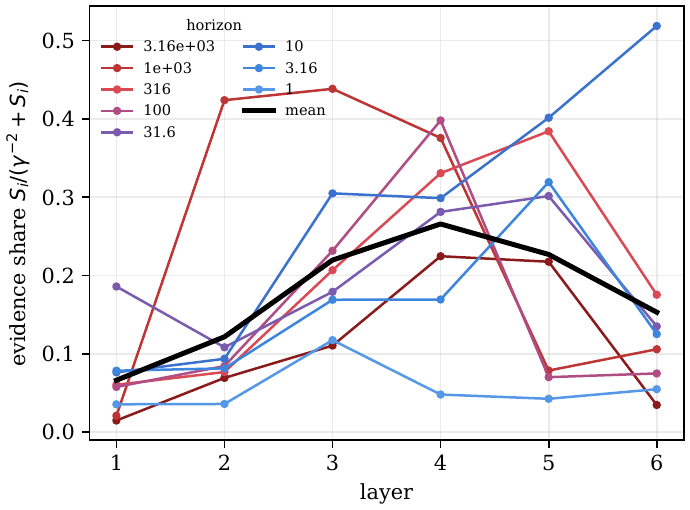}
  \caption{evidence share $S_i/(\gamma^{-2}+S_i)$}
  \label{fig:dka_depth_share}
\end{subfigure}
\caption{\textbf{(a)} The null slot's weight $c_h$ and \textbf{(b)} the evidence share $S_i/(\gamma^{-2}+S_i)$ across the six layers of the trained model. Both panels show one line per head from the slowest (dark red) to the fastest (blue). The solid black lines indicate the geometric mean over heads in (a) and the arithmetic mean in (b). The share is the training-time batch mean, averaged over the last 1000 steps.}
\label{fig:dka_depth}
\end{figure}

Across depth (Figure~\ref{fig:dka_depth}), the null slot's weight falls by two orders of magnitude from the first layer to the fourth (geometric mean over
heads 0.18 to 0.002) and partly recovers in the last two, and the evidence
share mirrors it, rising from 0.07 to 0.27: the middle layers, which hold the
induction heads described below, draw most on the context, while the first and
last layers lean on the prior. The fastest head mostly abstains: its share
stays below 0.12 in every layer.

\end{document}